# A Dual-TransUNet Deep Learning Framework for Multi-Source Precipitation Merging and Improving Seasonal and Extreme Estimates

Yuchen Ye[1,2], Zixuan Qi[4], Shixuan Li[4], Wei Qi[4], Yanpeng Cai[4], Chaoxia Yuan[1,3,*]

[1] State Key Laboratory of Climate System Prediction and Risk Management (CPRM) /Key Laboratory of Meteorological Disaster, Ministry of Education (KLME) /Collaborative Innovation Center on Forecast and Evaluation of Meteorological Disasters (CIC-FEMD), Nanjing University of Information Science and Technology, Nanjing 210044, China

[2] School of Atmospheric Sciences, Nanjing University of Information Science and Technology, Nanjing 210044, China

[3] School of Future Technology, Nanjing University of Information Science and Technology, Nanjing 210044, China

[4] Guangdong Basic Research Center of Excellence for Ecological Security and Green Development, Guangdong Provincial Key Laboratory of Water Quality Improvement and Ecological Restoration for Watersheds, School of Ecology, Environment and Resources, Guangdong University of Technology, Guangzhou, 510006, China

Corresponding author: Chaoxia Yuan (chaoxia.yuan@nuist.edu.cn)

**Highlights**

- Proposed a two-stage deep learning framework (DDL-MSPMF) that fuses six MSPs with ERA5 physical predictors for daily precipitation correction.

- Achieved the best seasonal correction with a TransUNet – TransUNet hybrid, outperforming single-regressor and previously proposed hybrid baselines.

- Enhanced heavy-rain detection (ETS improvements over eastern China) and explained key drivers using SHAP, highlighting the role of classifier outputs and physical predictors.

**Abstract**

Multi-source precipitation products (MSPs) from satellite retrievals and reanalysis are widely used for hydroclimatic monitoring, yet spatially heterogeneous biases and limited skill for extremes still constrain their hydrologic utility. Here we develop a dual-stage TransUNet-based multi-source precipitation merging framework (DDL-MSPMF) that integrates six MSPs with four ERA5 near-surface physical predictors. A first-stage classifier estimates daily precipitation occurrence probability, and a second-stage regressor fuses the classifier outputs together with all predictors to estimate daily precipitation amount at 0.25° resolution over China for 2001–2020. Benchmarking against multiple deep learning and hybrid baselines shows that the TransUNet – TransUNet configuration yields the best seasonal performance (R = 0.75; RMSE = 2.70 mm/day) and improves robustness relative to a single-regressor setting. For heavy precipitation (>25 mm/day), DDL-MSPMF increases equitable threat scores across most regions of eastern China and better reproduces the spatial pattern of the July 2021 Zhengzhou rainstorm, indicating enhanced extreme-event detection beyond seasonal-mean corrections. Independent evaluation over the Qinghai–Tibet Plateau using TPHiPr further supports its applicability in data-scarce regions. SHAP analysis highlights the importance of precipitation occurrence probabilities and surface pressure, providing physically interpretable diagnostics. The proposed framework offers a scalable and explainable approach for precipitation fusion and extreme-event assessment.

# 1. Introduction

Since the 20th century, significant changes have occurred in the spatiotemporal distribution characteristics of global precipitation. These changes are manifested not only in the increase of annual precipitation totals (S Li et al., 2024a), the rising frequency and intensity of extreme precipitation events (Green et al., 2025; Myhre et al., 2019), but also in the uneven seasonal distribution (Feng et al., 2013; Konapala et al., 2020) and the intensification of regional drought/flood extremes (L Z Chen et al., 2025b; Pizzorni et al., 2024; Y P Wu et al., 2024). These changes have been further exacerbated against the backdrop of global warming, posing greater demands on ecosystem security, water resource regulation, and flood disaster management (Eekhout et al., 2018; Tabari, 2020). Therefore, enhancing the spatiotemporal accuracy of precipitation observation and monitoring is of great significance for different countries and regions to formulate scientific climate risk response and disaster reduction strategies.

In recent years, alongside traditional ground-based precipitation meteorological stations, various novel observation methods such as satellite remote sensing retrievals, weather radar detection, and reanalysis data have continuously emerged, offering new possibilities for obtaining high spatiotemporal resolution precipitation data (X Wang et al., 2023). However, each single observation method inevitably has its limitations (Prein & Gobiet, 2016; Yu et al., 2020). For instance, ground stations are sparsely distributed in developing countries and data-scarce regions, making it difficult to comprehensively reflect regional precipitation distribution characteristics (Awadallah & Awadallah, 2013; Massari et al., 2020). Satellite observation products are susceptible to surface types, atmospheric conditions, and retrieval algorithms, leading to significant uncertainties in estimating extreme precipitation and complex terrain areas (Bartsotas et al., 2018; Maggioni et al., 2022). While reanalysis-based precipitation products can provide long-term time series data, their ability to reproduce extreme precipitation events is limited (Rhodes et al., 2014; X M Sun & Barros, 2010). To address the shortcomings of single data sources, multi-source precipitation products (MSPs) fusion and bias correction techniques have garnered widespread attention in recent years (Y Li et al., 2021; Nie et al., 2015; Pan et al., 2022). Existing research has shown that integrating multiple sources and employing efficient bias correction methods can significantly enhance the spatial coverage and accuracy performance of precipitation products (W Y Li et al., 2022b; Nan et al., 2024).

The accuracy of multi-source merged precipitation products largely depends on effective data fusion and bias correction methods (Assiri & Qureshi, 2022; W Y Li et al., 2022b). For a long time, traditional statistical approaches have been widely used for the fusion and correction of MSPs, including simple linear or weighted averaging (Y Xu et al., 2024), quantile mapping (Heo et al., 2019), cumulative distribution function (CDF) matching (Heo et al., 2019), and multiplicative shift (S S Qi et al., 2023). These methods improve product consistency across various temporal scales (monthly, seasonal, annual) by adjusting systematic biases between MSPs and ground observations. For instance, Bhatti et al. (2016) significantly reduced daily precipitation biases in satellite products like The Climate Prediction Center Morphing Technique Precipitation Climate Data Record (CMORPH) using a continuous sliding-window multiplicative shift approach, while Katiraie-Boroujerdy et al. (2020) enhanced The Precipitation Estimation from Remotely Sensed Information using Artificial Neural Networks Climate Data Record

(PERSIANN) performance at multiple temporal scales through quantile mapping and CDF matching. However, these methods commonly face limitations such as dependence on high-quality ground observations, overly strong assumptions about statistical relationship stability, and limited capability in capturing spatiotemporal variability and extreme events (Bhatti et al., 2016; Mendez et al., 2020).

In recent years, intelligent fusion methods such as machine learning and deep learning have demonstrated strong technological breakthroughs in the field of MSPs fusion and correction (Ardabili et al., 2020; Gavahi et al., 2023; X Z Xu et al., 2025; Zhang et al., 2022). Machine learning approaches represented by random forests, support vector machines, and gradient boosting trees can effectively uncover nonlinear relationships and complementarity among multi-source observations (Y Guo et al., 2012; Y M Zhao et al., 2022a; Y Y Zhao et al., 2022b). For instance, Nguyen et al. (2021) significantly improved the accuracy and stability of regional precipitation estimation by integrating multiple MSPs and ground observations using random forests, outperforming traditional statistical fusion methods. Furthermore, deep learning models, particularly neural networks such as UNet and Long Short-Term Memory (LSTM), have emerged as new hotspots in Earth system modeling and extreme weather event monitoring.(Le et al., 2023) conducted MSPs correction studies based on UNet neural networks and found that they outperformed traditional methods in capturing extreme rainfall, identifying spatial distributions, and modeling grid correlations. Yang et al. (2022) demonstrated the great potential of an improved LSTM network in correcting tropical MSPs biases. Despite these significant advancements, current research still faces several key scientific challenges. First, most existing deep learning fusion models rely solely on MSPs as input, neglecting the incorporation of meteorological physical information such as meteorological fields and model forecasts, which limits their ability to characterize precipitation processes under complex climatic drivers (F R Chen et al., 2024a; Gavahi et al., 2023). Recently, AI modeling methods incorporating meteorological physical predictors have shown unique advantages in meteorological forecasting (Carpenter, 2024; Mu et al., 2023; Zhou et al., 2024) but their application in MSPs bias correction remains underexplored. Secondly, mainstream fusion and correction methods mostly adopt a "mixed event" modeling approach, and further exploration is still warranted for refining different precipitation events. Existing studies have shown that MSPs exhibit significant deficiencies across different precipitation intensity levels (H Q Chen et al., 2023a). The latest hybrid models and event-specific modeling approaches (Lyu & Yong, 2024) provide new insights for addressing this scientific challenge. Therefore, developing fusion and correction techniques, incorporate multi-modal inputs, and are optimized for different precipitation event types is a core direction for enhancing the scientific value of MSPs.

In this study, we innovatively propose a Dual-TransUNet Deep Learning Multi-Source Precipitation Merging Framework (DDL-MSPMF). This framework aims to overcome the limitations of traditional multi-source merged precipitation products in terms of accuracy and interpretability, enabling a more refined and physically consistent characterization of spatiotemporal precipitation distribution. The DDL-MSPMF framework adopts a "dual-stage event-magnitude" deep learning hybrid modeling strategy. The first stage focuses on accurate discrimination of precipitation events, while the second stage performs regression estimation for specific precipitation amounts. Notably, we incorporate four key meteorological factors closely related to precipitation formation and evolution: surface pressure, 2 m air temperature, 2 m dew point temperature, and surface soil moisture at 0-7 cm into the multi-source data fusion process. This not only significantly enhances the model's ability to capture precipitation variations under complex meteorological scenarios but also improves the physical consistency and interpretability of the merged product. To ensure global scalability and applicability across multiple spatiotemporal scales, we selected six internationally recognized, high-resolution, and continuously updated MSPs datasets: CMORPH, the fifth-generation European Centre for Medium-Range Weather Forecasts reanalysis product (ERA5), the Global Precipitation Measurement (GPM), the Global Satellite Mapping of Precipitation (GSMAP), the Multi-Source Weighted-Ensemble Precipitation (MSWEP), and PERSIANN. Additionally, the CN05.1 precipitation product, constructed from 2,428 meteorological stations in China, and the Qinghai-Tibet Plateau high-resolution precipitation observation dataset were used as independent benchmarks to validate the generalization capability and accuracy improvement of our proposed fusion framework. In terms of model architecture, the first stage of DDL-MSPMF (precipitation event classification) systematically evaluated five representative models: XGBoost, convolutional neural networks (CNN)-Transformer, UNet, Transformer, and TransUNet. The second stage (precipitation amount regression) expanded to six models, including LSTM, CNN – Transformer, UNet, Transformer, TransUNet, and TransUNet_Direct. Through systematic screening and combination of model performances under different scenarios, six hybrid models were ultimately constructed, achieving optimal fusion corrections for both precipitation events and magnitudes. For evaluation, this study comprehensively assessed the multi-source merged precipitation product generated by DDL-MSPMF across three dimensions: accuracy in precipitation event detection, seasonal precipitation simulation, and extreme precipitation simulation capabilities, highlighting its applicability and scientific value across multiple spatiotemporal scales and complex climatic regions. The primary objectives of this study are: (a) To systematically compare the accuracy of DDL-MSPMF with existing mainstream multi-source fusion techniques in China, emphasizing its physical consistency and generalization capability; (b) To develop a continuously updatable and scalable high-precision multi-source merged precipitation product for China, providing foundational support for regional meteorological services and extreme event monitoring; (c) To explore the physical and statistical mechanisms underlying DDL-MSPMF's superiority and its potential for global expansion across different climatic regions, offering new insights for multimodal data fusion and intelligent meteorological observation.

## 2. Data

### 2.1 Data Summary

This study utilized six types of MSPs datasets, two types of observed precipitation datasets, and one type of meteorological physical predictor dataset. The temporal-spatial resolution, coverage, data access links, and providing institutions are detailed in Table S1.

### 2.2 Multi-Source Precipitation Dataset

CMORPH CDR provided by the National Oceanic and Atmospheric Administration (NOAA)/

National Centers for Environmental Information (NCEI), is a high-resolution global satellite precipitation product (Noaa National Centers for Environmental Information, 2019). This dataset has a spatial resolution of 0.25° and a temporal resolution of 1 hour, covering the entire globe from 1998 to the present. It performs well in China but tends to overestimate summer precipitation (Shen et al., 2020).

PERSIANN also released by NOAA for hydrological and climate research, offers long-term, high-resolution global precipitation observations (Ashouri et al., 2015). Its spatial resolution is 0.25°, with a daily temporal resolution, covering 60°S–60°N and 180°W–180°E from 1983 to 2021. PERSIANN effectively captures the spatiotemporal distribution of extreme precipitation in eastern China but performs relatively poorly in western China (Miao et al., 2015).

GPM developed by National Aeronautics and Space Administration (NASA), integrates multi-satellite observations and ground-based precipitation data to provide high-precision precipitation estimates (Huffman et al., 2024). It has a spatial resolution of 0.1°, a daily temporal resolution, and global coverage from 1998 to the present. In China, GPM generally outperforms the Tropical Rainfall Measuring Mission (TRMM), though its reliability is limited in winter and high-latitude regions (F R Chen & Li, 2016). In this study, the GPM dataset we used is officially named "TRMM (TMPA) Precipitation L3 day 0.25 degree × 0.25 degree V7 (TRMM_3B42 Daily 7)".

GSMAP developed by the Japan Aerospace Exploration Agency (JAXA), retrieves global

precipitation using multi-source satellite data (Kubota et al., 2020). The GSMAP dataset has a spatial resolution of 0.1°, a temporal resolution of 1 hour, and covers a spatial range from 60°S to 60°N and 180°W to 180°E, with a temporal coverage from 2000 to the present. This dataset effectively captures the spatial distribution and intensity of daily average precipitation, particularly performing well in mainland China during summer and in southeastern regions. However, it tends to underestimate total precipitation and is susceptible to interference from snow cover and vegetation (Z Q Chen et al., 2016). In this study, we utilized the Near Real-Time (NRT) version of GSMAP.

MSWEP merges multiple data sources to create a high spatiotemporal resolution global precipitation product (Beck et al., 2019). It has a spatial resolution of 0.1°, a 3-hour temporal resolution, and global coverage from 1979 to the present. In China, MSWEP performs well in temporal resolution but has limitations in spatial patterns and extreme precipitation representation (L J Li et al., 2022a).

ERA5, released by The European Centre for Medium-Range Weather Forecasts (ECMWF), integrates ground, satellite, upper-air, and radar observations, providing high spatiotemporal resolution precipitation data (Copernicus Climate Change Service & Climate Data Store, 2023). ERA5 has a spatial resolution of 0.25°, an hourly temporal resolution, and global coverage from 1940 to the present. It accurately reflects China's annual total precipitation but has certain limitations in simulating extreme precipitation events (X Y Lei et al., 2022b).

## 2.3 Precipitation Observation Data

This study employs the CN05.1 dataset (J Wu & Gao, 2013) as the primary precipitation observation benchmark. Compiled by the China Climate Research Center (CCRC), this dataset integrates observational data from over 2,400 meteorological stations across China, offering extensive spatial coverage and high spatial resolution. It has become one of the most widely used foundational datasets in China's hydro-meteorological field, extensively applied in regional climate studies and the evaluation of MSPs. CN05.1 performs exceptionally well in high-density observation areas, though it has certain limitations in complex terrains such as plateaus and mountainous regions. Its reliability as a benchmark observational dataset has been validated by numerous studies (Han et al., 2023). The dataset features a temporal resolution of 1 day and a spatial resolution of 0.25°, covering the period from 1961 to the present, with a spatial range of 69.75°E–140.25°E and 14.75°N–55.25°N. However, due to the sparse distribution of national benchmark meteorological stations in high-altitude regions, the CN05.1 dataset exhibits significant uncertainty in areas such as the Qinghai-Tibet Plateau. To address the observational gaps in this region, this study further incorporates the "Third Pole High-accuracy Precipitation dataset (TPHiPr)" released by the National Tibetan Plateau Data Center (TPDC). TPHiPr is derived from downscaled ERA5 reanalysis data and integrates observations from over 9,000 rainfall stations, effectively enhancing the spatial accuracy and reliability of precipitation data for the Qinghai-Tibet Plateau (Jiang et al., 2023). This dataset has a temporal resolution of 1 day and a spatial resolution of 1/30°, covering the period from 1979 to 2020, with a spatial range of 61.03°E–105.66°E and 25.73°N–41.36°N.

**2.4 Meteorological Factors Data**

This study employs the ERA5 hourly data on single levels from 1940 to present (Copernicus Climate Change Service & Climate Data Store, 2023) as meteorological physical factors in the MSPs fusion correction experiments. The selected variables include surface pressure, 2-meter air temperature, 2-meter dewpoint temperature, and 0-7 cm soil moisture. The ERA5 single-level dataset features a temporal resolution of 1 hour and a spatial resolution of 0.25°, with a temporal coverage from January 1, 1940, to the present, spanning global regions.

**2.5 Data Unification**

To standardize the study period and spatial scope, this paper selected data from 2001 to 2020, covering a research area of 18.0°–53.5°N and 73.5°–135.0°E. To enhance the readability of spatial representation, provincial administrative boundaries of China (Fig. S1a) and boundaries of China's nine major agricultural regions (Fig. S1b) were annotated on the maps. During data preprocessing, to maintain consistency in temporal and spatial resolution, all meteorological variables were first averaged on a daily basis, unifying the temporal resolution to 1 day. Subsequently, linear interpolation was employed to resample the spatial resolution, standardizing all data to a 0.25° spatial resolution. After linear interpolation, the data still maintains a high correlation with the original data (the correlation coefficients before and after interpolation are all above 0.99), preserving the characteristics of the original data. Each variable is standardized along the time dimension using z-score scaling (subtracting the mean and dividing by the standard deviation). We avoid min–max normalization because daily precipitation is strongly heavy-tailed and extremes may fall outside the historical range used to define the minimum/maximum. In such cases, min–max scaling can require extrapolation and/or clipping and tends to compress tail variability, which may degrade the correction of rare heavy-rain events (e.g., the 21 July 2021 Zhengzhou rainstorm). The model's output results are then uniformly destandardized along the time dimension (multiplying by the standard deviation and adding the mean) to restore the data's original scale. The data is split into training, validation, and test sets using sequential partitioning, with proportions of 0.64, 0.16, and 0.2 respectively. During the results presentation phase, all our outcomes are derived from the test set.

## 3. Methods

### 3.1 Dual-TransUNet Deep Learning Multi-Source Precipitation Merging Framework (DDL-MSPMF)

This study constructs a DDL-MSPMF that integrates classification and regression tasks for high-precision MSPs merging. Compared to existing "dual machine learning" precipitation merging methods (H J Lei et al., 2022a; Lyu & Yong, 2024), our framework achieves systematic improvements in model architecture, feature fusion, and the incorporation of meteorological physical forecast factors. It serves as a baseline model and is evaluated against various advanced deep learning architectures to comprehensively validate its superiority. The overall workflow is illustrated in Fig. 1 and consists of the following three steps:

(1) Classification Task Modeling: Using surface soil moisture, surface pressure, 2-meter temperature, 2-meter dew point temperature, and MSPs as inputs, multiple classifiers are trained to discriminate daily precipitation events. The classification labels are provided by CN05.1 observed precipitation data (threshold: 0.1 mm/day), and the binary cross-entropy loss function is employed to optimize the model, enhancing the ability to discern precipitation occurrence.

(2) Regression Task Modeling: The precipitation event probabilities filtered by the classifier, along with all meteorological and soil variables, are input into the regressor. The regression labels are provided by CN05.1 daily observed precipitation data, and the MSE loss function is used to optimize model parameters, achieving high-precision precipitation simulation. In all deep learning hybrid models, the classifier and regressor architectures are consistent. The baseline model adopts XGBoost – LSTM, while other models include UNet – UNet, Transformer – Transformer, TransUNet – TransUNet, and CNN – Transformer – CNN – Transformer.

(3) Comprehensive Evaluation and Physical Interpretation: Leveraging the TPHiPr dataset, the study systematically evaluates the precipitation simulation performance of MSPs based on different hybrid merging strategies in high-altitude and data-scarce regions. Extreme precipitation identification metrics such as ETS are used to specifically assess the models' ability to reproduce typical extreme events, such as the July 2021 Zhengzhou rainstorm. Additionally, the study deconstructs the precipitation simulation capabilities of each regressor model step-by-step and quantitatively evaluates the performance enhancement of regressors due to different classifier outputs using the Critical Success Index (CSI). Furthermore, we progressively deconstruct the precipitation simulation capabilities of various regression models. We quantitatively evaluate the performance enhancement effects of different classifier outputs on regression models. This systematic comparison examines the performance of single deep learning models, traditional machine learning models, and dual deep learning hybrid models in MSPs fusion tasks, highlighting the hybrid deep learning model's advantages in simulation accuracy and generalization ability. Finally, the SHapley Additive exPlanations (SHAP) method is employed to analyze the interpretability of the optimal model, identifying and quantifying the relative contributions of different precipitation products and meteorological physical forecast factors in the MSPs merging process.

This framework significantly improves the spatial consistency of MSPs merging, extreme event identification capability, and the interpretability of physical mechanisms, demonstrating stronger generalization and robustness in complex terrain and plateau extreme precipitation simulations.

All models in this paper were trained on a small server equipped with an AMD Ryzen Threadripper PRO 5975WX 32-Core CPU, 512GB of RAM, and an RTX 5090 GPU. The programming language used was Python, with the frontend developed using Jupyter Notebook under the Anaconda software environment. TensorFlow 2.10 served as the library package source for all deep learning models.

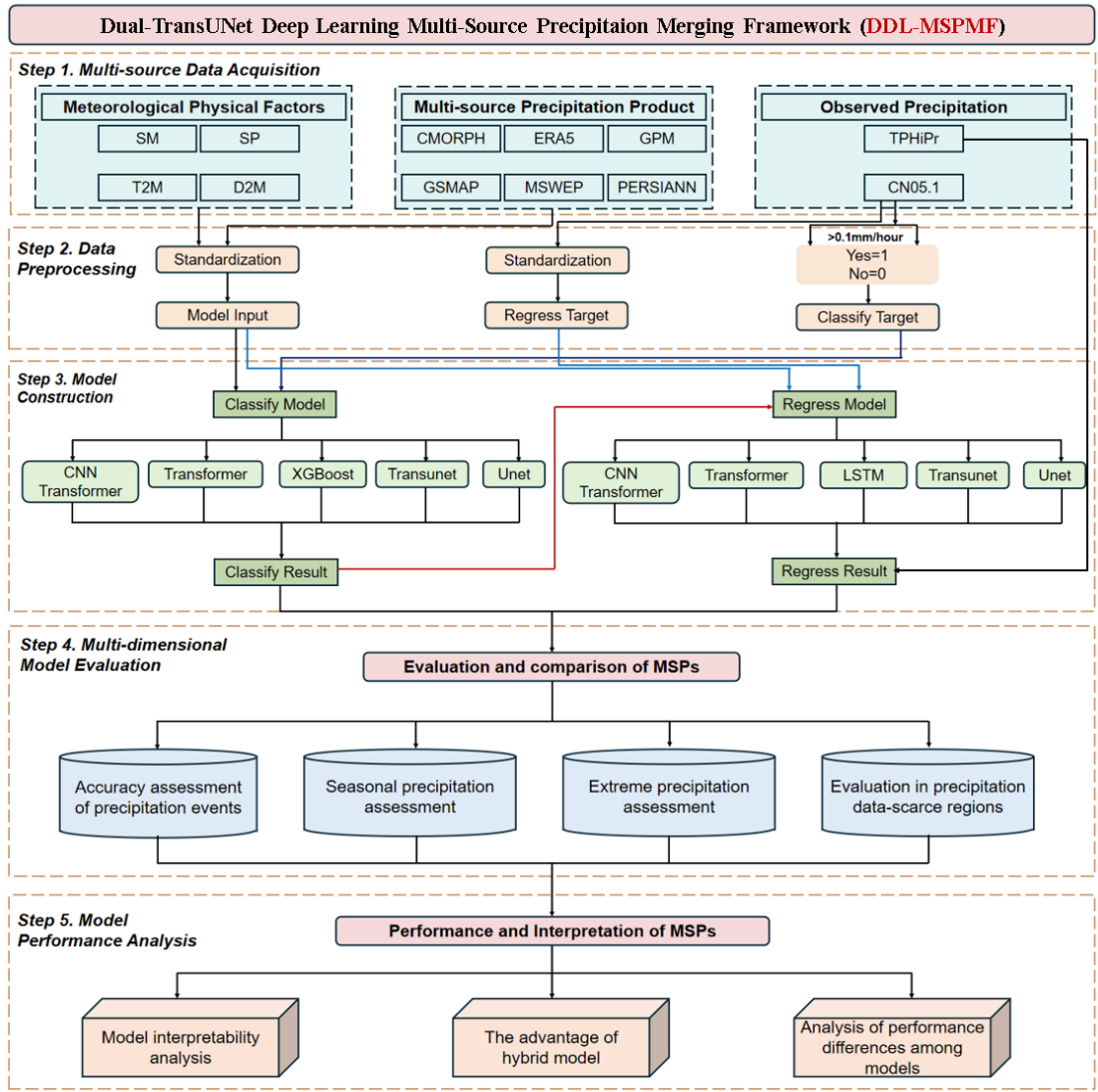

**Fig. 1.** Dual-TransUNet Deep Learning Multi-Source Precipitation Merging Framework Diagram (DDL-MSPMF).

## 3.2 Machine Learning Architecture and Model Details

**3.2.1 XGBoost**

XGBoost is an ensemble decision tree algorithm based on the Boosting framework (T Chen & Guestrin, 2016), which has been widely used in classification and regression tasks in recent years as an efficient model. The decision tree algorithm was first proposed by Quinlan (1979) and has evolved through multiple iterations, leading to various improved versions. As a typical ensemble learning method, the fundamental concept of Boosting consists of the following three steps: (1) First, construct an initial decision tree and calculate its objective function value and prediction residuals; (2) Then, introduce a new decision tree to fit the residual portion not yet captured by the current model, while ensuring the new model reduces the overall objective function; (3) Repeat the above process until the predefined termination conditions are met. In this study, we employ XGBoost as the baseline classifier. XGBoost takes two-dimensional input (batch, feature) and outputs one-dimensional (batch). We merge the lat and lon dimensions of the original input (batch, lat, lon, feature) into the batch dimension, then reshape the batch dimension back into three dimensions (batch, lat, lon) for the XGBoost output. The specific model parameters include: a learning rate set to 0.1, a maximum tree depth of 50, a minimum leaf node sample size of 100, and a maximum number of feature splits (number of bins) of 256.

## 3.3 Deep Learning Architectures and Model Details

**3.3.1 Long Short-Term Memory (LSTM)**

LSTM network was proposed by Hochreiter and Schmidhuber in 1997 and was initially applied to natural language processing (NLP) tasks (Hochreiter & Schmidhuber, 1997). Compared to traditional recurrent neural networks (RNNs), LSTM effectively mitigates the vanishing gradient problem by introducing memory cells and gating mechanisms, while also exhibiting stronger long-term dependency modeling capabilities. It can more flexibly control the forgetting, updating, and output of information, making it suitable for complex time-series modeling tasks. In this study, LSTM is employed for precipitation classification and regression modeling. In terms of network architecture, the LSTM layer is followed by a fully connected layer (with 1024 neurons). The classifier network uses a Sigmoid activation function at the output, whereas the regressor network does not employ an activation function; The final output dimension of the model is the same as the original data's batch*lat*lon*feature. The structure of the LSTM network is illustrated in Fig. S2. The timesteps are set to 5, indicating that the model uses data from the previous five time steps (i.e., t−4 to t) to correct precipitation conditions at the current time step (t). LSTM can only accept three-dimensional input (batch, timesteps, feature) and output three-dimensional/two-dimensional data (since our model is not a seq2seq model, the output is two-dimensional). Therefore, we merge the latitude and longitude dimensions of the input data into the batch dimension, and then reshape the batch dimension back into three dimensions (batch, lat, lon) during output. The reason we did not merge lat and lon into the feature dimension is that our input data has a grid size of 143*247 (latitude * longitude). If merged into the feature dimension, the feature dimension size would become 388,531 (143*247*11), which is nearly impossible to implement on a single RTX 5090 GPU or even any mainstream single GPU (such as NVIDIA A100). Such a large dimension size would consume an enormous amount of video memory. During model training, the initial learning rate is set to 0.001, the optimizer is Adam, the loss function is MSE, and the total number of training epochs is 500. The rationale for selecting a learning rate of 0.001 and 500 epochs in this study is that an appropriate learning rate is crucial for network training. If the learning rate is too high, the network parameters struggle to approach the global optimum; conversely, if it is too low, the required training epochs increase substantially (Wilson & Martinez, 2001). Therefore, we chose a learning rate of 0.001, which ensures both rapid network convergence and proximity to the global optimum (reducing the rate to 0.0001 would increase the optimal training iterations tenfold, incurring excessive computational costs). With this learning rate, the optimal number of epochs was found to be 242. However, to guarantee complete convergence, we trained the network for 500 epochs. In our experiments, training for 500 epochs did not result in significant overfitting compared to the 242 epoch (detailed results are available in the Jupyter notebook provided in the GitHub repository under Data Availability Statement).

### 3.3.2 UNet

UNet (Ronneberger et al., 2015) is a variant architecture based on CNN (Lecun et al., 1998), composed of convolutional layers, pooling layers, and upsampling modules (Gu et al., 2018). Fig. S3 illustrates its main structure: after every two convolutional operations, a max pooling is performed, repeating this "convolution-convolution-pooling" sequence four times, followed by two convolutional operations before entering the symmetric upsampling path. The upsampling section similarly follows the sequence of "upsampling-convolution-convolution", with each upsampling layer connected via skip connections to the feature maps of corresponding resolution from the encoder path, followed by convolutional operations to fuse the features. The network starts with 64 convolutional kernels, doubling after each pooling and halving after each upsampling, with the final output layer using 1 kernel. All convolutional layers except the output layer employ the Tanh activation function; for regression tasks, the output layer uses no activation function, while for classification tasks, the Sigmoid function is applied for final activation. The UNet network accepts 4-dimensional input (batch, lat, lon, feature), while the first three dimensions of the output remain exactly the same as the input. We don't need to perform any additional processing on the data. The model is trained using the Adam optimizer with an initial learning rate of 0.001 over 500 epochs, employing binary cross-entropy (BCE) as the classifier loss function and mean squared error (MSE) as the regressor loss function.

### 3.3.3 Transformer

The Transformer model was initially proposed by Vaswani et al. (2017) as a classic NLP model. Compared to traditional RNNs (Hopfield, 1982), the Transformer enables parallel processing of input sequences through its positional encoding mechanism, whereas RNNs require sequential iteration due to their temporal dependency and cannot perform parallel computation. The overall architecture of the Transformer consists of an encoder and a decoder: the encoder extracts input features, while the decoder performs sequence prediction, generating the next time step's input based on the current output. However, in this study, the input comprises combinations of multiple meteorological predictors across multiple time steps, and the output is only the target variable (precipitation) at a single time step. Therefore, only the Transformer's encoder is employed for feature extraction, followed by a fully connected layer to produce the output, resulting in a concise overall structure. The fully connected layer is configured with 1,024 neurons and uses the Tanh activation function. The internal fully connected layers within the encoder are also set to 1,024 neurons to maintain consistent feature dimensionality processing. For time-series modeling, the Transformer's timestep length is set to 5, meaning data from time steps -4 to 0 are used to correct precipitation at time step 0. The overall structure is illustrated in Fig. S2, resembling the LSTM model architecture but with the sequence processing unit replaced by the Transformer encoder module. For classification tasks, a Sigmoid activation function is added to the Transformer's output, while no activation function is used for regression tasks; The final output dimension of the model is the same as the original data's batch*lat*lon*feature. Similar to LSTM, the Transformer accepts three-dimensional input (batch, timesteps, feature) and produces output of either three or two dimensions. We still incorporate the lat and lon dimensions of the input data into the batch, ultimately reshaping the model output into three dimensions (batch, lat, lon). During training, the model employs the Adam optimizer with an initial learning rate of 0.001, runs for 500 epochs, employing BCE as the classifier loss function and MSE as the regressor loss function.

**3.3.4 CNN – Transformer**

CNN was first proposed by Lecun *et al.* (1998) and is capable of effectively extracting spatial features from samples. Among its components, the convolutional layer performs sliding calculations on input feature maps using convolution kernels to achieve local feature extraction, serving as the core element for spatial modeling. Subsequently, Karpov et al. (2020) introduced the CNN – Transformer model, which combines CNN's spatial feature extraction capability with Transformer's advantage in temporal sequence modeling, resulting in stronger spatiotemporal information representation. This study adopts the CNN – Transformer architecture illustrated in Fig. S4: spatial features are first extracted through multiple convolutional and pooling layers, followed by a Transformer Encoder module to model temporal dependencies, and finally, the correction results are output via a fully connected layer. The model's timestep is set to 5, meaning it uses input data from time steps -4 to 0 to correct precipitation at time step 0. The specific parameter settings are as follows: model depth (N) is 2, convolution kernel size is 3×3, and the number of kernels sequentially increases to 16, 32, 64, and 64. The Tanh activation function is applied to intermediate layers, while the output layer uses the Sigmoid function for classification tasks and no activation function for regression tasks; The final output dimension of the model is the same as the original data's batch*lat*lon*feature. The CNN – Transformer accepts four-dimensional input (batch, lat, lon, feature), yet its output is two-dimensional. We reshape the output of the CNN – Transformer into (batch, lat, lon) to maintain dimensional consistency with the CN05.1 precipitation ground truth. The model is trained using the Adam optimizer with an initial learning rate of 0.001 for 500 epochs, employing BCE as the classifier loss function and MSE as the regressor loss function.

### 3.3.5 TransUNet

TransUNet is a hybrid architecture proposed by J N Chen et al. (2021) in 2021, which embeds Transformer modules into the encoder of the traditional UNet to enhance its capability for modeling global features. Compared to the standard UNet, its overall workflow remains consistent (see Fig. S3), with the key difference being the introduction of a Transformer encoder layer after the last MaxPooling layer to improve the model's ability to capture long-range dependencies. In this study, the classifier part of TransUNet employs a Sigmoid activation function at its final layer, while the regressor part does not include any activation function at its output. Both TransUNet and

UNet share the same input and output dimensions, accepting four-dimensional inputs (batch, lat, lon, feature) and producing three-dimensional outputs (batch, lat, lon). No additional processing is required for TransUNet's inputs and outputs. The training strategy is identical to that of UNet, with an initial learning rate set to 0.001, using the Adam optimizer, BCE as the classifier loss function and MSE as the regressor loss function, trained for a total of 500 epochs.

**3.4 Model Interpretability**

SHAP was proposed by Lundberg et al. in 2017 (Lundberg & Lee, 2017)and has been widely applied in fields such as precipitation prediction and fusion correction (C Chen et al., 2025a; W Li et al., 2024b; Lyu & Yong, 2025). Assuming the prediction model is f(x), with the input predictor feature set M, SHAP employs an additive explanatory model to approximate the prediction for a single sample x:

$$g(z) = \varphi_0 + \sum_{i=1}^{M} \varphi_i z_i \quad (1)$$

Here, z indicates whether a feature is included, represented as a binary vector (taking values 0 or 1), $\varphi_i$ is the contribution (SHAP value) of the i-th feature to the prediction for the sample, and $\varphi_0$ denotes the baseline, which is the average value of the background field (typically, n randomly selected samples serve as the background field). As for

$$\varphi_i(x) = \frac{1}{B}\sum_{b=1}^{B} c_i(x, x^{(b)}) \qquad (2)$$

where B represents the background sample set, and

$$c_i(x, x^{(b)}) = m_i(x, x^{(b)}) \cdot (x_i - x_i^{(b)}) \qquad (3)$$

Here, $m_i$ represents the multiplier obtained through hierarchical propagation in the network, satisfying additivity. Simply put, SHAP examines the influence of a single predictor $x_i$ relative to the background field B on the predicted value y. By averaging all dimensions of the SHAP value (sample dimension, spatial dimension) except the feature dimension for x and y, we can calculate the impact of each feature of the predictor x on each output channel of the predicted value y.

**3.5 Evaluation Metrics for Multi-source Precipitation Products**

**3.5.1 Seasonal Precipitation Assessment**

In the seasonal precipitation assessment, this study employs the Pearson Correlation Coefficient (R) and RMSE as core evaluation metrics. These two indicators are the most widely used statistical measures for evaluating the seasonal performance of MSPs (Y Y Zhao et al., 2022b). The R measures the degree of linear correlation between observed and simulated sequences, reflecting their consistency in temporal variation trends, with values ranging from -1 to 1, values closer to 1 indicate stronger correlation. The RMSE quantifies the magnitude of differences between the two sequences at each corresponding time point, where smaller values indicate closer agreement between simulated results and observations, reflecting the model's accuracy performance. The calculation formulas for these two metrics are as follows:

$$R = \frac{\sum(x_i-\bar{x})(y_i-\bar{y})}{\sqrt{\sum(x_i-\bar{x})^2 \sum(y_i-\bar{y})^2}} \quad (4)$$

Here, $x_i$ and $y_i$ represent each value in the two sequences, while $\bar{x}$ and $\bar{y}$ denote the time-dimensional means of the two sequences. An r close to 1 indicates a perfect positive correlation between the two sequences, an r close to 0 suggests linear independence, and an r close to -1 signifies a perfect negative correlation.

$$RMSE = \sqrt{\frac{1}{mn}\sum_{i=1}^{m}\sum_{j=1}^{n}(x(i,j) - y(i,j))^2} \quad (5)$$

Here, $x(i,j)$ and $y(i,j)$ represent the values at each grid point in two three-dimensional sequences, while m and n denote the number of grid points along the longitude and latitude dimensions. The RMSE ranges from 0 to +∞, with 0 being the best and +∞ being the worst.

**3.5.2 Precipitation Event Accuracy Assessment**

In the performance evaluation of MSPs, accurately identifying precipitation events is a crucial aspect of assessing product effectiveness (Y Deng et al., 2024). To this end, this study introduces the CSI as a supplementary metric to quantitatively evaluate the precipitation event identification capabilities of different products (Y Li et al., 2021). The CSI comprehensively considers both the Probability of Detection (POD) and the False Alarm Ratio (FAR), providing a more holistic reflection of prediction reliability and effectiveness. Its calculation formula is as follows:

$$CSI = \frac{H}{H+F+M} \tag{6}$$

Here, H represents the number of hits, F represents the number of false alarms, and M represents the number of misses. The CSI ranges from 0 to 1, with 0 being the worst and 1 being the best.

**3.5.3 Assessment of Extreme Precipitation Events**

In the evaluation of extreme precipitation events, the Equitable Threat Score (ETS) is widely adopted as one of the key metrics for assessing a model's ability to identify extreme events (C-C Wang, 2014). By removing the influence of random hits in its calculation, the ETS provides a more equitable reflection of a model's actual predictive capability for extreme precipitation events, making it particularly suitable for identifying low-frequency but high-impact precipitation scenarios. The specific calculation formula is as follows:

$$ETS = \frac{H - H_{random}}{H + F + M - H_{random}} \tag{7}$$

Here, H represents the number of hits, F represents the number of false alarms, and M represents the number of misses.

$$H_{random} = \frac{(H+F) \cdot (H+M)}{N} \tag{8}$$

Here, N is the total number of samples. The ETS value ranges from -1/3 to 1, where a negative value indicates worse performance than random prediction, a positive value indicates better performance than random prediction, and 1 represents perfect prediction.

## 4. Result

### 4.1 Performance of DDL-MSPMF in Simulating Seasonal Precipitation Over China

Fig. S5 displays the R (Fig. S5a) and RMSE (Fig. S5b) between different MSPs and the CN05.1 precipitation data. In the results, the names of precipitation products have been simplified: C_R represents CMORPH precipitation, E_P represents ERA5 precipitation, G_P represents GPM precipitation, GS_P represents GSMAP precipitation, M_P represents MSWEP precipitation, and P_C represents PERSIANN precipitation. The agreement between the precipitation data and CN05.1 is relatively low. The highest R is only 0.556 (ERA5), while the lowest drops to 0.321 (CMO). The RMSE ranges from 4.107 mm/day to 6.249 mm/day, indicating significant deviations between the precipitation products and observational data across most times and grid points. The excessively low R and high RMSE values further suggest that the reliability of these MSPs data is generally low in most cases. Further analysis was conducted on the spatial correlation and RMSE distribution between the MSPs ensemble mean and CN05.1 precipitation. Fig. S6a shows that the R between MSPs and CN05.1 are more pronounced in eastern China, particularly in the Loess Plateau, Huang-Huai-Hai Plain, Southern China, and the Middle-lower Yangtze River, where the coefficients approach 0.6. However, in the northern arid and semiarid regions, southeastern Sichuan, and Qinghai, the R are significantly lower. Notably, minimal values appear in eastern and southwestern Xinjiang, southeastern Sichuan, and northwestern Guizhou, with some areas even below 0.1. The RMSE of the MSPs ensemble mean relative to CN05.1 precipitation exhibits a spatial pattern of lower values in the northwest and higher values in the southeast (Fig. S6b).

Fig. 2 presents the R (Fig. 2a) and RMSE (Fig. 2b) between the MSPs fusion correction results based on the DDL-MSPMF framework and CN05.1 precipitation for different deep learning models. Table S2 displays the specific numerical results and total runtime corresponding to Fig.2. In Fig. 2, the names of the deep-learning models are simplified as follows: C_T indicates that both the classifier and regressor use CNN – Transformer; UNet indicates that both the classifier and regressor use UNet; T_U indicates that both the classifier and regressor use TransUNet; TF indicates that both the classifier and regressor use Transformer; LSTM indicates the classifier uses XGBoost and the regressor uses LSTM (Baseline model); TU_D represents TransUNet_Direct, where all data is directly input into the regressor without a classifier. From Fig. 2a and Fig. 2b, it is evident that deep learning models significantly improve the R and RMSE performance of MSPs. Except for CNN – Transformer and XGBoost – LSTM, all other models outperform individual MSPs in both R and RMSE. The remaining models achieve R values ranging from

0.659 to 0.751 and RMSE values between 2.695 mm/day and 3.025 mm/day. The CNN – Transformer model exhibits lower R and higher RMSE, likely due to its convolution and pooling operations, which may lead to the loss of image detail information. In contrast, UNet and TransUNet maintain consistent input-output image sizes and employ skip connections to preserve details, while the Transformer network avoids convolution and pooling altogether by integrating latitude (lat) and longitude (lon) directly into feature dimensions, thus preventing detail loss. The XGBoost – LSTM combination, serving as a baseline model, shows decent R (0.626) but suffers from extreme RMSE deviation (19.865 mm/day), far exceeding that of individual MSPs. Among all models, TransUNet performs the best, achieving an R of 0.751 and an RMSE of 2.695 mm/day. Additionally, the Hybrid model provides a slight performance improvement. The TransUNet – TransUNet model, which first conducts precipitation classification before quantitative correction, shows clear enhancements over TransUNet_Direct (R: 0.703, RMSE: 2.940). The TransUNet model incorporating the self-attention mechanism outperforms the UNet model without self-attention.

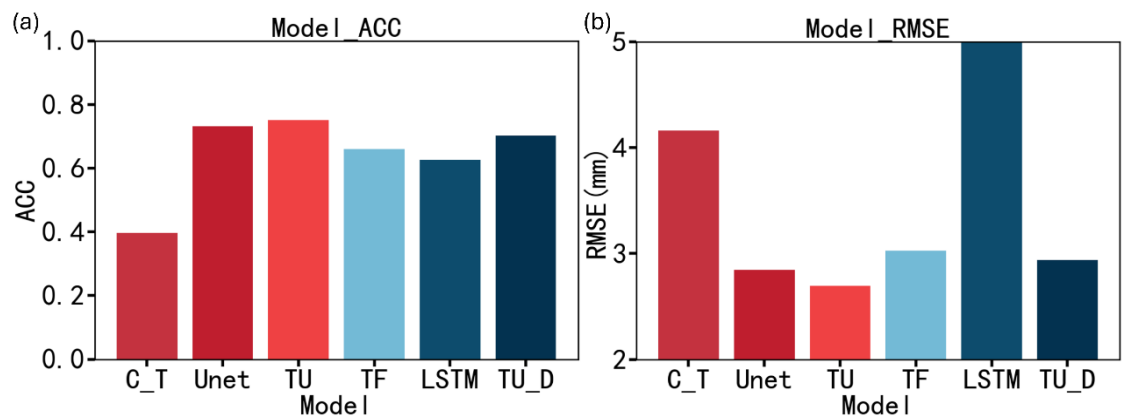

**Fig. 2.** Spatial average R (a) / RMSE (b) between the MSPs data fusion correction results of different deep learning models and the CN05.1 observed precipitation; Since the RMSE of LSTM is 19.865 mm/day, significantly higher than other models, it was not fully displayed.

Fig. 3 displays spatial maps of the R and RMSE between precipitation results from deep learning models based on different DDL-MSPMF framework outputs and CN05.1 precipitation, as well as the RMSE differences between the ensemble mean precipitation results of MSPs and CN05.1 precipitation. From Fig. 3a-f, it can be observed that, except for the CNN – Transformer model, all other deep learning models show improvements in R compared to MSPs. Specifically, the CNN – Transformer model only exhibits improvements over the Qinghai-Tibet Plateau, northeastern China, and parts of northwestern China, while in eastern China, its R are lower than those of MSPs, dropping by up to around 0.3. The LSTM model shows limited improvement in Xinjiang, the Northern Arid and Semiarid Region, the Loess Plateau, and Southern China, with almost no improvement in Xinjiang. The Transformer model has a spatial distribution similar to that of the LSTM model but demonstrates greater improvement in R over the northern Qinghai-Tibet Plateau, southern Sichuan, and northern Guizhou. The TransUNet-based models exhibit nearly universal improvement in R across China, with the most significant enhancement observed over the northern Qinghai-Tibet Plateau. The UNet and TransUNet_Direct models show negligible improvement in northern Henan, possibly due to the impact of the Zhengzhou 7.20 extreme rainfall event (which will be discussed in later sections), whereas the TransUNet model shows notable improvement in this region. From Fig. 3g-l, it can be seen that, except for XGBoost – LSTM, all other deep learning models improve the RMSE of the MSPs ensemble mean results. The XGBoost – LSTM model exhibits high RMSE values primarily concentrated in the Yangtze

River Basin and Southern China, with the maximum RMSE exceeding the MSPs ensemble mean by up to 140 mm/day. For the high RMSE in northern Xinjiang, all deep learning models except XGBoost – LSTM show improvement. The CNN – Transformer performs unsatisfactorily in improving the RMSE of the MSPs ensemble mean, particularly in northern Henan and central Guangzhou, where RMSE remains higher than the MSPs ensemble mean. The remaining models: Transformer, TransUNet, UNet, and UNet_Direct demonstrate significant improvement over the MSPs ensemble mean in most regions, with the degree of improvement following a pattern of lesser enhancement in the northwest and greater enhancement in the southeast (except for northern Xinjiang, where improvement is notable). In Southern China, these four models reduce the RMSE of the MSPs ensemble mean by up to 8.5 mm/day. In summary, among the deep learning models in the DDL-MSPMF framework, except for CNN – Transformer (which has lower R than MSPs) and XGBoost – LSTM (which has higher RMSE), all other models outperform MSPs ensemble mean, with the TransUNet model performing the best.

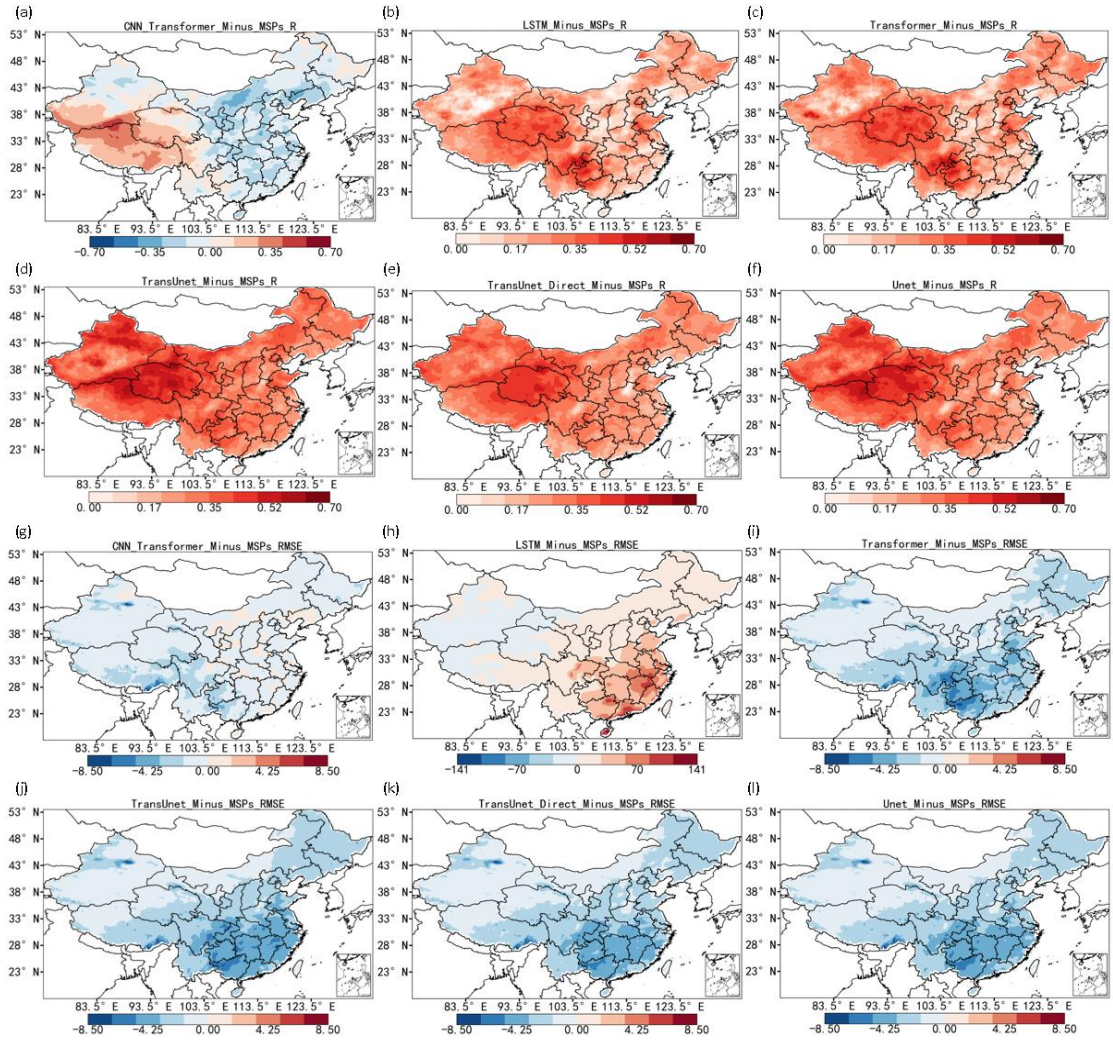

**Fig. 3.** The spatial maps of R (a-f) / RMSE (g-l) between different deep learning models' corrected MSPs data fusion results and CN05.1 observed precipitation, minus the R (a-f) / RMSE (g-l) between the MSPs ensemble mean results and CN05.1 observed precipitation.

**4.2 Performance of DDL-MSPMF in Simulating Seasonal Precipitation in Data-Scarce Regions of China**

However, the CN05.1 dataset is derived from station data, and the station distribution on the Qinghai-Tibet Plateau is relatively sparse, which may lead to inaccuracies in precipitation data for this region. Therefore, the reliability of DDL-MSPMF's precipitation improvements on the

Qinghai-Tibet Plateau warrants further investigation. To address this issue, we introduced the TPHiPr dataset, which provides more comprehensive observational data for the Qinghai-Tibet Plateau, effectively compensating for the scarcity of station data in CN05.1 for this area. Fig. S7 displays the R (Fig. S7a) and RMSE (Fig. S7b) between MSPs and the TPHiPr dataset. As shown in Fig. S7, the R between MSPs and TPHiPr exhibit a pattern of lower values in the west and higher values in the east of the Qinghai-Tibet Plateau, with the lowest values in the west approaching 0 and the highest values in the east and south reaching up to 0.8. Meanwhile, the RMSE shows larger values in the southern and eastern parts of the Qinghai-Tibet Plateau, with the highest RMSE in southern Tibet, approaching 16 mm/day. Fig. 4 illustrates the improvements in R and RMSE achieved by deep learning models compared to the ensemble mean results of MSPs. From Fig. 4, it can be observed that the three models: CNN – Transformer, LSTM, and Transformer primarily enhance the performance of MSPs in the central and western parts of the Qinghai-Tibet Plateau. However, their performance deteriorates in some eastern regions, even leading to a decline in precipitation forecasting accuracy. In contrast, the three UNet-based models (TransUNet, UNet, and TransUNet_Direct) demonstrate improvements across nearly the entire Qinghai-Tibet Plateau, with only minor degradation in very few areas. Among them, TransUNet_Direct exhibits the most significant improvements. The UNet-based models achieve a maximum improvement in R of up to 0.6, highlighting their superior performance in the Qinghai-Tibet Plateau region. In terms of RMSE, all models except LSTM show improvements over MSPs across the entire Qinghai-Tibet Plateau, with the maximum improvement reaching 6.5 mm/day. This further validates the effectiveness of UNet-based models in this region, proving that deep learning models can significantly enhance MSPs results even in areas with sparse station coverage, demonstrating their exceptional model performance.

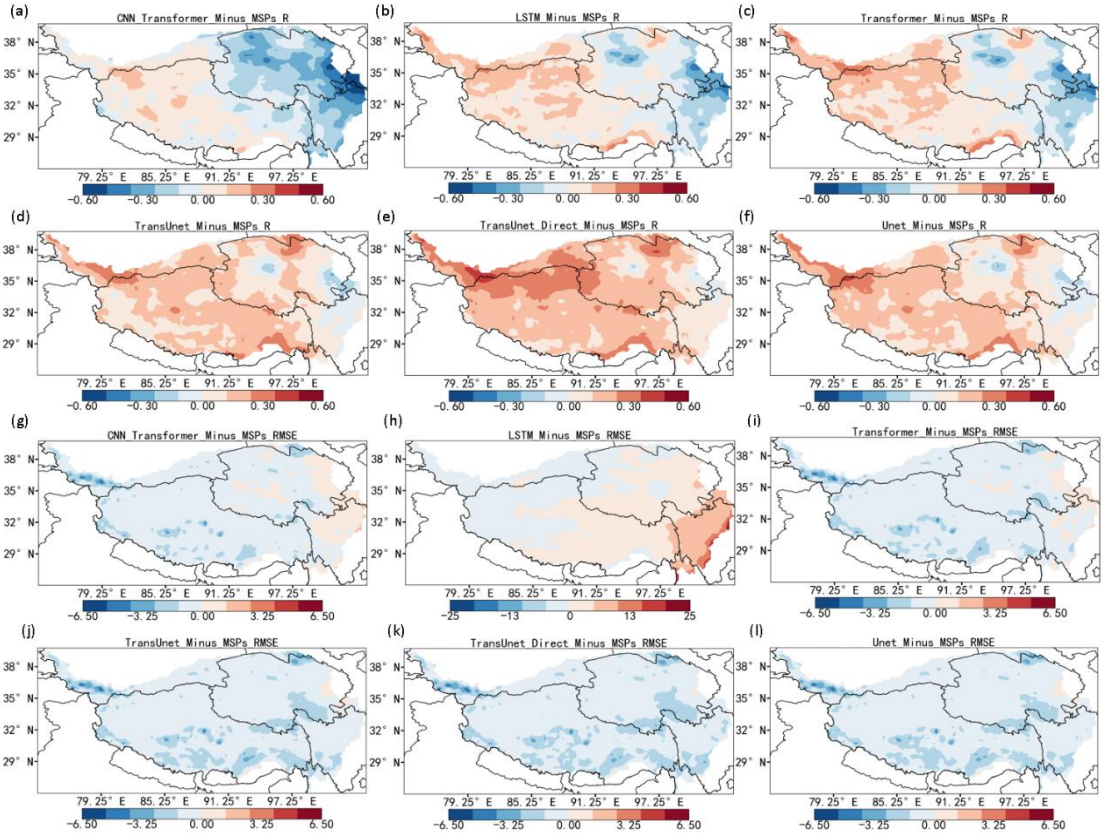

**Fig. 4.** The spatial maps of the R (a-f) / RMSE (g-l) between the MSPs data fusion correction results from different deep learning models and the TPHiPr observed precipitation, minus the R (a-f) / RMSE (g-l) between the MSPs ensemble mean results and the TPHiPr observed precipitation.

**4.3 Performance of DDL-MSPMF in Multi-source Precipitation Fusion Correction During Extreme Precipitation Events**

In the previous section, we demonstrated the ability of DDL-MSPMF to improve seasonal precipitation in MSPs. However, for precipitation, heavy rainfall events often lead to significant economic losses and casualties, making them a key focus of our attention (Kotz et al., 2022; Newman & Noy, 2023; Tabari, 2020). According to the GB/T 28592-2012 standard established by the Standardization Administration of China in 2012, we selected heavy rain and rainstorm events with daily precipitation exceeding 25 mm/day and calculated the ETS of different deep learning models in MSPs and DDL-MSPMF to evaluate the performance of DDL-MSPMF in capturing heavy precipitation events. Fig. S8 shows the spatial distribution of ETS for identifying heavy precipitation in the ensemble mean of MSPs. Due to the rarity of heavy precipitation events in the Northern arid and semiarid region, this analysis focuses on southeastern China. From Fig. S8, it can be observed that MSPs exhibit good identification capability for heavy precipitation events in Southern China and Sichuan, with relatively high ETS values. However, their identification ability is weaker in the Yunnan-Guizhou Plateau, Northeast China Plain, and parts of Inner Mongolia, as reflected by lower ETS values. Overall, the ETS of MSPs is relatively low, averaging only 0.36. According to the criteria of Owens & Hewson (2018), an ETS value greater than 0.3 in short-term forecasts is considered good, and MSPs meet this standard in only a few regions. Fig. 5 displays the difference in ETS for identifying heavy precipitation events between different deep learning models in DDL-MSPMF and the ensemble mean of MSPs. From Fig. 5, it can be seen that, except for CNN – Transformer, all other models show improvements in ETS compared to MSPs. For the LSTM model, the ETS improvement is most significant in the Yunnan-Guizhou Plateau and the northern part of the Northeast China Plain, with a maximum increase exceeding 0.3. The Transformer models show improvements in eastern China, but their ETS decreases compared to MSPs in parts of Inner Mongolia, the Yunnan-Guizhou Plateau, and the western Loess Plateau.

The results of UNet and TransUNet_Direct are similar, with both showing significant ETS improvements across most regions of China, except for minor declines in small parts of Inner Mongolia. Notably, TransUNet outperforms TransUNet_Direct in terms of ETS improvement in central Inner Mongolia and the southern Northeast China Plain, as well as in other regions of eastern China. Overall, the deep learning models in DDL-MSPMF, except for CNN – Transformer, all improve upon the ETS of MSPs, particularly the LSTM and TransUNet models. These two models can increase the ETS to above 0.3, with some regions even exceeding 0.5. This result demonstrates the reliability of DDL-MSPMF in identifying heavy precipitation events.

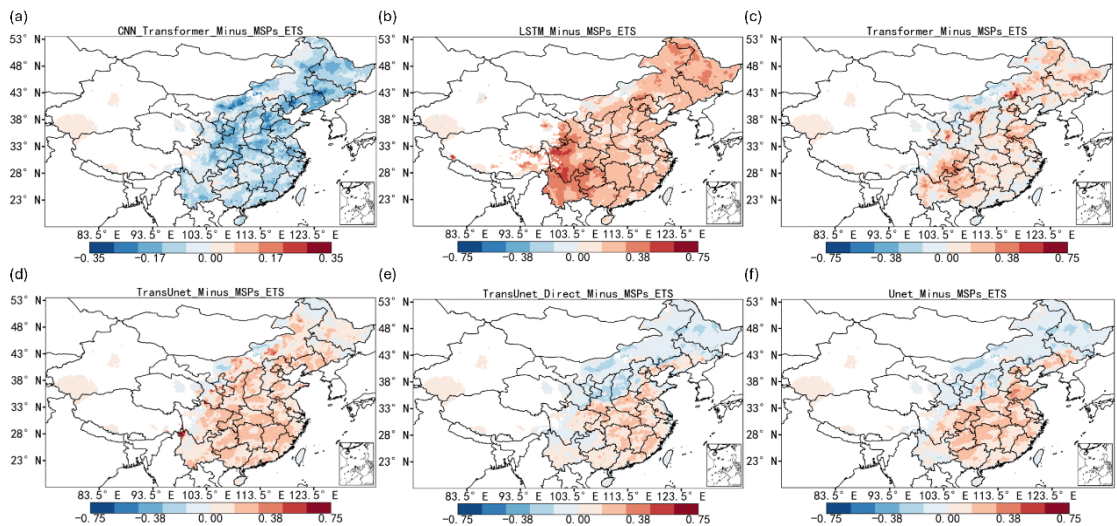

**Fig. 5.** Spatial map of ETS for identifying heavy precipitation events by different deep learning models with MSPs data fusion correction minus the ETS by MSPs ensemble mean for identifying heavy precipitation events; The white areas within China indicate regions without extreme precipitation events where daily rainfall exceeds 25mm/day.

Taking a specific extreme precipitation event occurring in China as an example, Henan Province experienced an exceptionally severe extreme precipitation event from July 17 to July 23, 2021, which triggered serious flooding (Zhai & Lee, 2023). We analyzed the performance of MSPs in capturing the "7.21" Zhengzhou rainstorm event. Fig. S9 displays the spatial distribution of the CN05.1 precipitation data on July 21, 2021, showing that extreme precipitation areas were primarily concentrated in northern and central Henan. In addition to these regions, significant precipitation zones also appeared in southwestern Guangdong and southern Guangxi, with intensities reaching up to 50 mm/day. We focused on evaluating MSPs in these two extreme/high-precipitation regions (marked by red boxes in Fig. S9). Fig. S10 illustrates the spatial distribution of precipitation from various MSPs on July 21, 2021. CMO_RPH's extreme precipitation zone mainly fell near the border between Henan and Hebei, missing the extreme zone in central Henan, while precipitation in southern Guangxi was also overestimated, with overall magnitudes significantly higher than observations (the maximum value exceeded observations by nearly 80 mm/day). ERA5_PRE captured the high-precipitation zones in southwestern Guangdong and southern Guangxi, but its extreme precipitation area was located at the junction of Henan, Hebei, and Shanxi, missing the extreme zone in central Henan, with overall magnitudes also overestimated (the maximum value exceeded observations by nearly 50 mm/day). GPM_PRE failed to capture precipitation in southwestern Guangdong, overestimated precipitation in southern Guangxi, and placed the extreme precipitation zone at the border of Henan and Hebei, also missing the extreme precipitation zone in central Henan, although its maximum precipitation magnitude was closer to observations. GSMAP_PRE's precipitation area was smaller, with overestimated intensity in southern Guangxi. The extreme precipitation zone in Henan was split into two parts—one in northern Henan and another in southern Henan—unlike the spatially continuous dual extreme precipitation centers in observations, and its maximum precipitation magnitude exceeded observations by about 90 mm/day. MSWEP_PRE also missed the extreme precipitation center in central Henan, while PERSIANN_CDR_PRE captured both extreme precipitation centers but with a maximum precipitation magnitude only half of observations, along with false precipitation zones in Tibet, Yunnan, and northeastern China. Overall, MSPs either failed to capture the extreme precipitation zone in central Henan or exhibited significant discrepancies in precipitation magnitudes compared to CN05.1 data.

Fig. 6 shows the spatial distribution of precipitation from different deep learning models in DDL-MSPMF on July 21, 2021. The results from CNN – Transformer were nearly unusable, with precipitation patterns and magnitudes inconsistent with observations. The LSTM model's precipitation pattern closely matched observations, but its magnitude (especially in southern Guangzhou) was severely overestimated, reaching nearly 800 mm/day—far exceeding historical daily precipitation records for Guangzhou—leading to higher RMSE for the LSTM model. The Transformer model's extreme precipitation centers aligned well with observations, with precipitation in southern Guangzhou, northern Henan, and central Henan matching observations in both pattern and magnitude, though the shape of its extreme precipitation centers was elliptical, differing from the elongated shape in observations. The TransUNet model's precipitation pattern in Henan was closest to observations, with extreme precipitation centers in both northern and central Henan connected in an elongated shape, and its maximum precipitation magnitude matched observations. However, it produced some false precipitation signals in Sichuan, Yunnan, and

Hainan. The TransUNet_Direct and UNet models showed similar precipitation distributions and magnitudes, with their extreme precipitation centers in northern and central Henan somewhat matching observations but covering larger areas. They also exhibited false precipitation in Yunnan, Sichuan, Hainan, and Xinjiang, with maximum precipitation magnitudes much lower than observations. The TransUNet model performed best in Henan, closely matching observations, and successfully captured the precipitation band in southern Guangzhou. Although false precipitation zones remained in Sichuan and southern Yunnan, their extent was smaller compared to TransUNet_Direct, and false precipitation in Tibet and Xinjiang was almost entirely eliminated. This improvement validates the effectiveness of the Hybrid model—the front-end classifier successfully corrected low-value false precipitation to no precipitation, reducing false precipitation area, while adjusting the magnitude of extreme precipitation to better align with observations.

In summary, among the deep learning models in DDL-MSPMF, TransUNet and Transformer were able to closely match the precipitation patterns in Fig. S9. Although these models slightly underestimated extreme precipitation, such biases could be corrected and improved using methods like CDF matching (Brocca et al., 2011; W F Guo et al., 2022; Reichle & Koster, 2004).

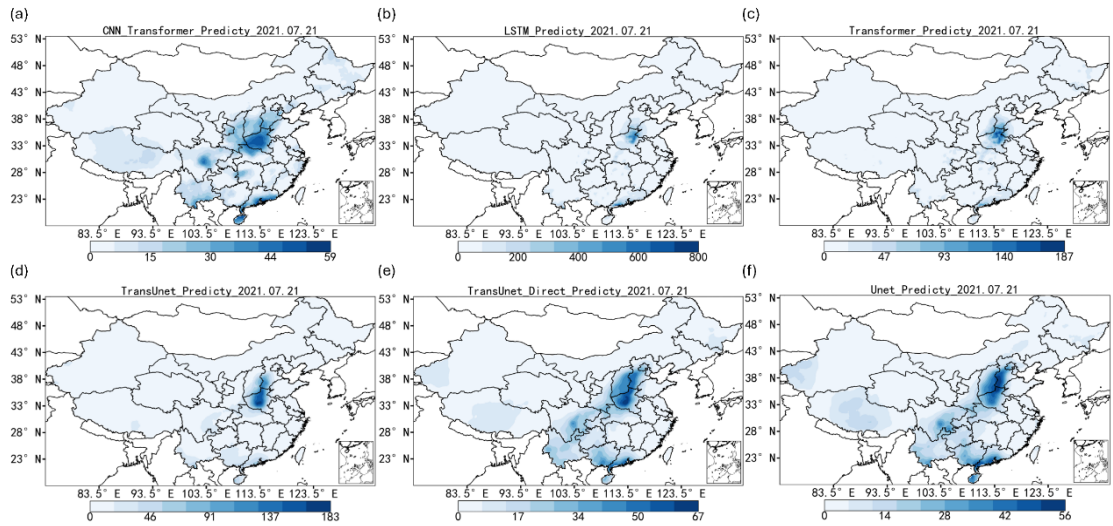

**Fig. 6.** The different deep learning models with MSPs data fusion correction on the spatial distribution of rainfall in China on July 21, 2021; The unit of precipitation is mm/day.

## 5 Discussion

**5.1 The superiority of the TransUNet – TransUNet Hybrid Model in MSP fusion**

In recent years, there has been no shortage of work on hybrid models for MSPs fusion correction (H J Lei et al., 2022a; Lyu & Yong, 2024; Senocak et al., 2023), with many studies employing UNet-based model for this purpose (Q Deng et al., 2023; Gao et al., 2022; D Wang et al., 2024). However, research utilizing the TransUNet – TransUNet hybrid model for MSPs fusion correction remains quite rare. In this section, we will discuss the superiority of the TransUNet – TransUNet hybrid model over previous approaches in MSPs fusion correction from two perspectives.

First, we will examine why the TransUNet – TransUNet hybrid model outperforms traditional machine learning hybrid models (such as RandomForest – RandomForest or XGBoost – LSTM). This study has already presented substantial results demonstrating the superiority of the TransUNet – TransUNet hybrid model over traditional XGBoost – LSTM hybrid models in terms of outcomes (Figs. 2-6). Our TransUNet -TransUNet hybrid model achieved a 0.12 improvement in correlation coefficient and reduced RMSE over 17mm/day compared to the XGBoost – LSTM hybrid model. Yet, why does the TransUNet – TransUNet hybrid model perform better than the conventional XGBoost – LSTM model? In fact, it's quite straightforward to understand from a theoretical perspective: TransUNet is a model in the field of CV (J N Chen et al., 2021), while LSTM belongs to the domain of NLP (Hochreiter & Schmidhuber, 1997). For our MSPs fusion correction, it aligns more closely with computer vision because the input data is three-dimensional (batch, lat, lon), where the relationships between adjacent grid points are highly correlated. For example, if a grid point at a certain time has heavy or torrential rain, its neighboring grid points are also highly likely to experience significant precipitation, rather than no rain or light rain. TransUNet, with its use of convolutions, can easily capture the relationships between adjacent grid points. On the other hand, LSTM is more focused on capturing correlations within time series data. However, in the correction task, since we already know the MSPs values for the current time step (though they may contain errors or inaccuracies), temporal information becomes less critical than spatial features (Gavahi et al., 2023; H J Lei et al., 2022a). This is also evident from our experimental results, where all NLP models (Transformer, LSTM) underperformed compared to CV models (TransUNet, UNet). Although LSTM can also integrate latitude and longitude into the feature dimensions and establish connections between spatial grid points through fully connected layers, our experiments show that this approach is too costly. Models built this way cannot run on single 32-GB GPU due to out-of-memory errors. However, for RF or XGBoost models, they are even less capable of capturing spatial features. Similar to LSTM, their performance is significantly inferior to the TransUNet model (Malik et al., 2021). Of course, in recent years, there have also been some attempts to embed LSTM models into UNet-based model to simultaneously capture spatiotemporal features (Yuan et al., 2024), which should also be considered in our future MSPs fusion and correction work. At the same time, we also found that TransUNet outperforms UNet in almost all prediction results, which is quite easy to explain. This is because UNet only uses convolutional layers as spatial feature extractors, focusing more on local spatial feature extraction, whereas TransUNet introduces Transformer self-attention into both the encoder and decoder, enabling it to capture global spatial features (J Chen et al., 2024b).

Second, compared to a single TransUNet regressor, our model achieves a maximum Pearson correlation coefficient increase of 0.0481 and reduces the root mean square error by 0.2447 mm/day. We hope to explain why the TransUNet – TransUNet hybrid model performs better than the single TransUNet regression model (TransUNet_Direct Model). In other words, we need to demonstrate the crucial role of the classifier in the hybrid model. Although preliminary experimental results indicate that the performance gap between the Hybrid model and non-Hybrid models (such as TransUNet and TransUNet_Direct) is not significant, this raises a question: Does the classifier play a substantial role in the fusion of MSPs? Can it be removed from the experiments? To address this, Fig. 7 displays the CSI spatial distribution of classifiers from different deep learning models in DDL-MSPMF for identifying precipitation events. Fig. 7 reveals that deep learning model classifiers exhibit strong recognition capabilities in southwestern China (particularly the South China region), with CSI values exceeding 0.6 across all models. Specifically, the CNN – Transformer model shows lower CSI values (approaching 0) in the Huang-Huai-Hai Plain, Northeast China Plain, Northern Arid and Semiarid Region. This primarily occurs because the CNN – Transformer has a higher false alarm rate (FAR) in these areas—meaning it incorrectly identifies non-precipitation events as precipitation in these relatively dry regions, resulting in lower CSI.

The Transformer model demonstrates improved CSI compared to CNN – Transformer in Northern Arid and Semiarid Region, the Huang-Huai-Hai Plain, and the Northeast China Plain, while its performance in other regions remains similar to CNN – Transformer. UNet performs worse than Transformer in Northern Arid and Semiarid Region and the eastern part of Northeast China Plain. In contrast, Xgboost shows almost no performance gap compared to Transformer. Among all models, TransUNet achieves the highest CSI, with a spatial average CSI of 0.5771 (CNN – Transformer: 0.1625, Xgboost: 0.5311, Transformer: 0.5513, UNet: 0.4564). This indicates that hybrid model classifiers (except CNN – Transformer) provide a solid foundation for subsequent regressor correction.

This study did not employ more classification categories because increasing the number of classes would reduce both CSI and accuracy—feeding poorly performing classifier results into regressors would degrade their performance. We attempted dividing classifiers into six groups (0.0–0.1 mm/day, 0.1–9.9 mm/day, 9.9–24.9 mm/day, 24.9–49.9 mm/day, 49.9–99.9 mm/day, >99 mm/day), but this multi-classifier performed significantly worse than the binary classifier. When using this classifier's output as regressor input, the correlation coefficient decreased by 0.0363 compared to the binary classifier (detailed experimental procedures are available on the GitHub repository provided in the Data Availability Statement).

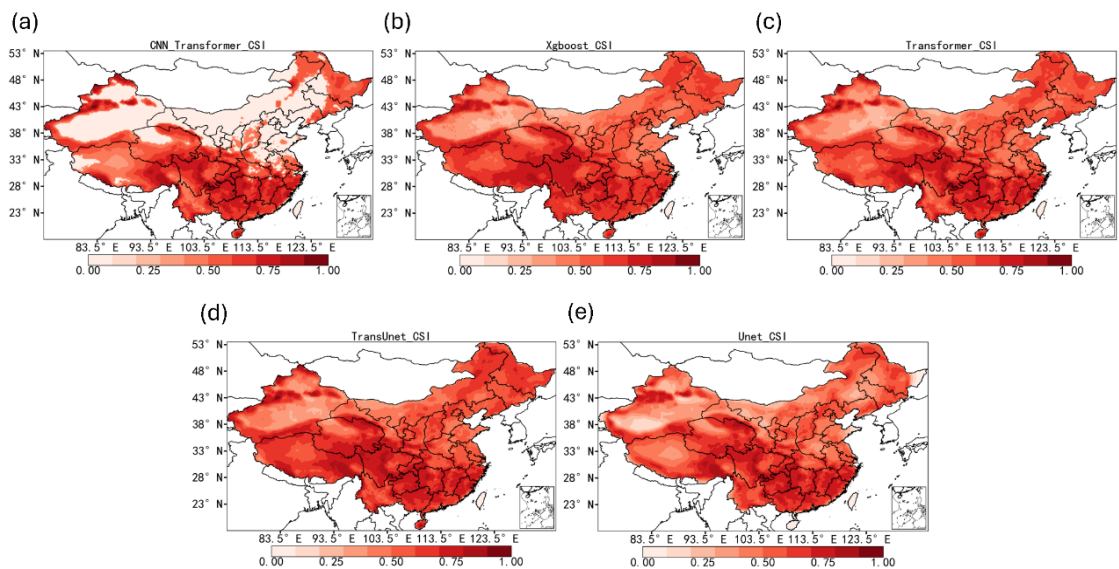

**Fig. 7.** Spatial distribution of CSI for identifying heavy precipitation events by different deep learning models using MSPs data fusion correction results.

Fig. 7 only demonstrates the superior performance of the classifier in the deep learning model but does not reveal its specific contribution to the correction of the regressor. To thoroughly analyze the role of the classifier, we employed the SHAP method to calculate both the correction magnitude (Fig. 8) and relative contribution (Fig. S11) of input values to output values in the

TransUNet model. Fig. 8 reveals that the classifier (CLASS) ranks fifth in its contribution to precipitation fusion correction. Notably, high classifier values (when equal to 1) exhibit a positive correction effect on precipitation, while low values (when equal to 0) show a negative correction effect—consistent with our expectations. In Fig. S11, the classifier's contribution accounts for approximately 6%, which, while not dominant, remains non-negligible: removing the classifier would degrade model performance. Its contribution value is comparable to P_C in multi-source precipitation data and even exceeds that of C_R and E_P.

Additionally, Fig. 8 and Fig. S11 shows that the contribution of SM is relatively low (only 0.22%), as precipitation typically influences soil moisture rather than the reverse (J Sun et al., 2025). Local evaporation driven by soil moisture accounts for only a minor portion of the moisture source for precipitation, with the primary source being non-local moisture transport (Van Der Ent & Savenije, 2011). This non-local moisture transport is associated with SP (L Guo et al., 2018), which, among meteorological and physical factors, ranks highest with a contribution of 10.96% in Fig. S11. Generally, surface pressure indirectly affects precipitation by influencing atmospheric vertical motion: low pressure promotes air convergence and uplift, enhancing moisture transport and condensation to form precipitation, while high pressure leads to air divergence and subsidence, increasing moisture dispersion and inhibiting condensation, thereby suppressing precipitation (You & Ting, 2021). This aligns with the conclusion in Fig. 8: positive SP exerts a negative correction on model output while negative SP shows positive correction. Meanwhile, Fig. 8 also shows that both 2-meter dew point temperature and 2-meter air temperature exhibit a positive correlation with precipitation, which is largely consistent with previous findings. This is because when near-surface dew point temperatures are low (indicating drier conditions), dry air tends to suppress the development of deep convection/precipitation (Ali et al., 2018). Conversely, when moisture availability is not limited or can be replenished through transport, rising 2-meter air temperatures increase both saturation vapor pressure and precipitable water, thereby enhancing extreme precipitation events as temperatures rise (Allen & Ingram, 2002).

Among MSPs factors, G_P and GS_P exhibit relatively high contributions (both exceeding 20%), indicating their greater importance compared to other MSPs factors, which aligns with previous research findings (F R Chen & Li, 2016; Z Q Chen et al., 2016). It also verifies the advantage of deep learning models: the ability to automatically assign weights to various factors, giving lower weights to factors weakly correlated with correction results. All multi-source precipitation data show that positive values have a positive correction on the model output, while negative values have a negative correction, demonstrating the robustness of the data we selected.

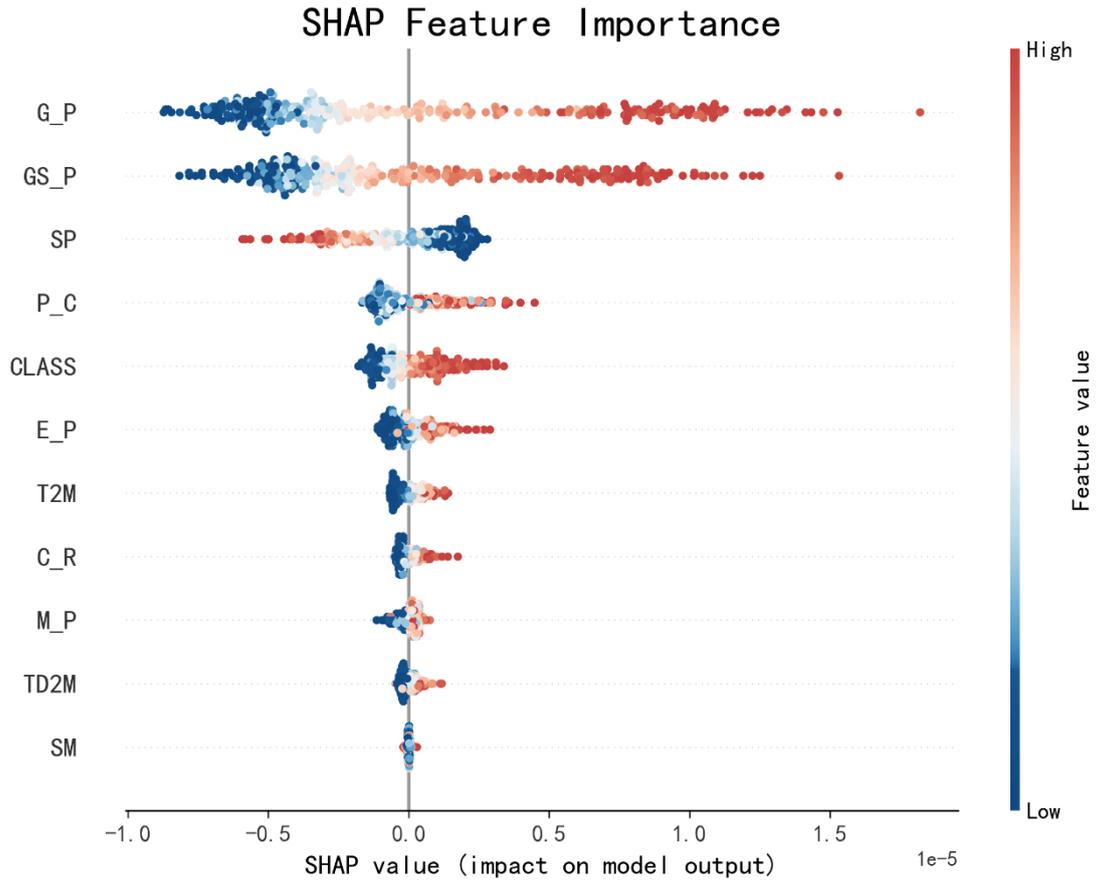

**Fig. 8.** Distribution of SHAP values for forecast factors in the TransUNet model. The plot reveals the directionality and magnitude of feature impacts on the fused precipitation correction.

To analyze the impact of the Hybrid model compared to a single regressor model on the fusion correction results of MSPs, we compared the spatial averages of the corrected values from the TransUNet and TransUNet_Direct models with the true spatial averages from the CN05.1 dataset, using scatter plots to illustrate the classifier's influence on the models' overall corrections. Fig. 9 presents the comparative results: Fig. 9a shows the TransUNet model's corrected values on the horizontal axis and the CN05.1 true values on the vertical axis, with the black diagonal line indicating perfect agreement between corrected and observed values. Theoretically, if all points lie on the diagonal, the model's correction is entirely accurate; points above the diagonal indicate underestimation, while those below indicate overestimation. The color represents point density, with red shades indicating more overlapping points. The red line is a polynomial fit to the scatter points—if it coincides perfectly with the black line, it proves the model's predicted mean matches observations exactly. Fig.9a reveals that the hybrid model performs well in correcting light precipitation, with near-perfect alignment in low-value regions. For heavy precipitation, there is slight underestimation (the red line lies above the black diagonal), but overall, the red trend line closely follows the diagonal without significant deviation.

In contrast, Fig. 9b demonstrates that the TransUNet_Direct model exhibits more pronounced underestimation for heavy precipitation: when spatially averaged true precipitation exceeds 5 mm/day, the model consistently underestimates (all points lie above the diagonal with no overestimation), and the red line visibly diverges from the diagonal, confirming severe precipitation underestimation—consistent with the conclusions from Fig. 6. For light precipitation, however, the model shows slight overestimation (the dense red region in Fig. 9b lies below the black line). Since the only difference between the TransUNet_Direct and TransUNet models is the absence of a classifier, the underestimation of heavy precipitation and overestimation of light precipitation can be attributed to the lack of a classifier, reaffirming its indispensable role in the hybrid model.

In summary, we have validated the superiority of the TransUNet – TransUNet hybrid model:

compared to traditional XGBoost – LSTM and RandomForest – RandomForest methods, it better captures the spatial features of MSPs data, achieving superior fusion correction. Relative to the conventional UNet model, TransUNet exhibits stronger spatial feature extraction capabilities. Compared to using the TransUNet regression model alone, the hybrid model, through the incorporation of a high-CSI classifier, significantly mitigates the tendency of the standalone TransUNet regression approach to underestimate heavy precipitation and overestimate light precipitation.

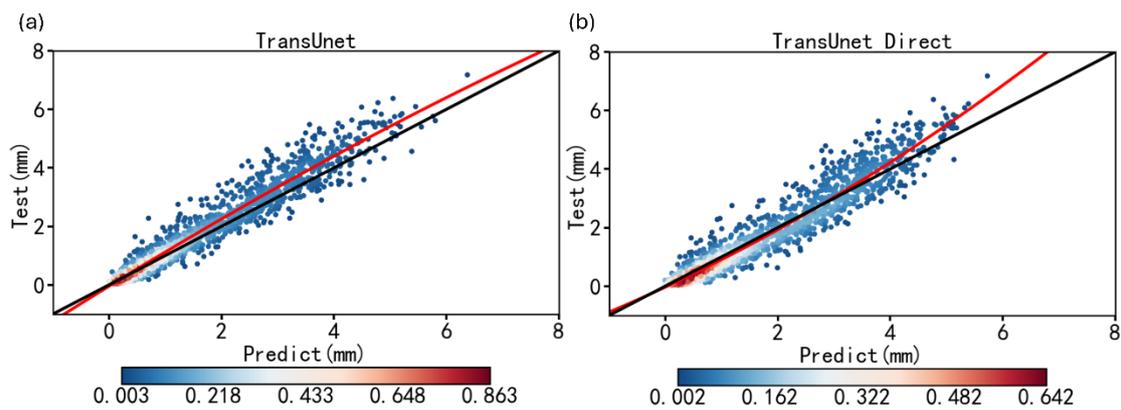

**Fig. 9.** Scatter plot comparing the spatially averaged results of MSPs fusion correction by TransUNet (a) and TransUNet_Direct (b) with the spatially averaged precipitation data from CN05.1; The red line represents the model fitting effect, the black diagonal line indicates the optimal fitting effect, and the closer the two are, the better; the color represents the scatter density.

**5.2 The Application Prospects of DDL-MSPMF in Future Multimodal Data Fusion**

The paper proposes the DDL-MSPMF framework and combines it with the Hybrid Model for the fusion correction of MSPs, achieving certain results. However, this study still has several limitations in the application of the Hybrid Model. Firstly, the research does not incorporate other important data such as radar precipitation data and numerical model precipitation data as inputs for the Hybrid Model. Yet, existing studies have shown that in MSPs fusion correction tasks, radar precipitation data (Y Qi et al., 2013; Sebastianelli et al., 2013) and the correction of numerical model precipitation data (Argüeso et al., 2013; Luo et al., 2022) have been extensively researched, widely applied, and yielded favorable outcomes. Therefore, future work could explore integrating radar precipitation data and numerical model precipitation data as inputs, combined with the Hybrid Model in DDL-MSPMF, to further enhance the model's forecasting performance. Additionally, recent research indicates that using multi-source data as front-end inputs and deep learning models as back-end outputs, coupled with concatenating both for data correction and prediction (the Fuxi-DA model), can significantly improve the stability and accuracy of precipitation forecasting (X Z Xu et al., 2025). Thus, in MSPs fusion correction, we could attempt to employ DDL-MSPMF as a front-end processing module and integrate it with deep learning models as the back-end, exploring the potential of this innovative approach to enhance precipitation correction performance. Lastly, it is worth noting that the deep learning model currently used in DDL-MSPMF is relatively basic, while meteorological large models and more sophisticated deep learning architectures have continued to emerge in recent years. Research demonstrates that, based on large-scale datasets and employing more complex deep learning network architectures, significant advantages over traditional numerical models can be achieved in weather forecasting and precipitation prediction (Bi et al., 2023; L Chen et al., 2023b). Future work should consider integrating more advanced deep learning architectures into the Hybrid Model of DDL-MSPMF, or even explore the application of meteorological large models, leveraging more meteorological data to further improve the effectiveness of MSPs fusion correction.

## 6. Conclusion

This study developed a Dual-TransUNet Deep Learning Multi-Source Precipitation Merging Framework (DDL-MSPMF) to produce bias-corrected 0.25° daily precipitation over China by jointly integrating six multi-source precipitation products (MSPs) with ERA5 near-surface physical predictors in a two-stage event–magnitude design. By explicitly separating "whether it rains" (classification) from "how much it rains" (regression), the framework mitigates spatially heterogeneous MSP biases, improves stability across regions with contrasting hydroclimates, and strengthens skill for heavy-rain diagnostics.

Comprehensive comparisons with TransUNet_Direct, UNet – UNet, and a conventional hybrid baseline (e.g., XGBoost – LSTM) demonstrate the advantages of the proposed framework. The main findings are:

(1) Most deep-learning configurations within DDL-MSPMF substantially improve seasonal precipitation skill relative to the raw MSPs in terms of correlation and RMSE. Among all tested combinations, the TransUNet – TransUNet hybrid achieves the best overall seasonal correction (R = 0.7512; RMSE = 2.6954 mm/day), outperforming both the single-regressor setting (TransUNet_Direct) and the traditional machine-learning hybrid baseline.

(2) For extreme precipitation, especially heavy rain and rainstorm events (daily precipitation > 25 mm/day), DDL-MSPMF increases equitable threat scores across most regions of eastern China, indicating that the framework improves event detection and spatial placement beyond seasonal-mean bias reduction.

(3) Mechanistically, the classifier is a necessary component for improving the subsequent regressor. TransUNet yields the highest CSI among tested classifiers (spatial-average CSI = 0.5771), and SHAP attribution indicates that the classifier output contributes 5.88% of the overall correction skill. Physical predictors are automatically reweighted by the model; surface pressure provides the largest contribution among the physical factors (10.96%), whereas soil moisture contributes marginally (0.22%). Removing the classifier leads to systematic underestimation of heavy precipitation and overestimation of light precipitation, degrading both the intensity distribution and spatial consistency.

Overall, DDL-MSPMF provides an accurate and interpretable pathway for multi-source precipitation fusion, with clear benefits for hydroclimatic monitoring and extreme-event assessment in China. Future work will focus on (i) incorporating additional observation and model-based inputs such as weather radar and numerical-weather-prediction precipitation, (ii) testing more advanced deep-learning architectures and meteorological foundation models within the DDL-MSPMF pipeline, and (iii) coupling the TransUNet-based framework with downscaling methods to generate higher spatiotemporal resolution precipitation products that better support climate-impact assessments, early warning of extremes, and water-resources management.

**Data Availability Statement**

The citation information and access links for the data used in this study are detailed in Chapter 2 and the supplementary materials. The MSPs datasets employed include: (1) The Climate Prediction Center Morphing Technique Precipitation Climate Data Record (CMORPH; https://www.ncei.noaa.gov/products/climate-data-records/precipitation-cmorph); (2) The Precipitation Estimation from Remotely Sensed Information using Artificial Neural Networks Climate Data Record (PERSIANN; https://climatedataguide.ucar.edu/climate-data/persiann-cdr-precipitation-estimation-remotely-sensed-information-using-artificial); (3) The Global Precipitation Measurement (GPM) Integrated Multi-satellitE Retrievals for GPM (IMERG) (https://gpm.nasa.gov/); (4) The Global Satellite Mapping of Precipitation (GSMAP; https://sharaku.eorc.jaxa.jp/GSMaP/guide.html); (5) Multi-Source Weighted-Ensemble Precipitation (MSWEP; https://www.gloh2o.org/mswep/) (6) European Centre for Medium-Range Weather Forecasts (ECMWF) Fifth Generation Reanalysis (ERA5) hourly data on single levels from 1940 to present (https://cds.climate.copernicus.eu/datasets/reanalysis-era5-single-levels). The precipitation observation datasets used include: (1)Third Pole High-accuracy Precipitation dataset (TPHiPr; https://data.tpdc.ac.cn/en/data/e45be858-bcb2-4fea-bd10-5c2662cb34a5); (2) CN05.1 (https://ccrc.iap.ac.cn/resource/detail?id=228). The meteorological physical predictors used are derived from European Centre for Medium-Range Weather Forecasts (ECMWF) Fifth Generation Reanalysis (ERA5) hourly data on single levels from 1940 to present (https://cds.climate.copernicus.eu/datasets/reanalysis-era5-single-levels).

Use the Anaconda software (https://www.anaconda.com/download), which includes Jupyter Notebook as the foundational programming software, using Python as the primary programming language. All fundamental deep learning modules are sourced from the open-source library TensorFlow (https://tensorflow.google.cn/).

The detailed information of the code used in this paper is as follows:

Software Name: DDL-MSPMF

Developer: Yuchen Ye

Contact: 492947833@qq.com

First Available Date: July 14, 2025

Required Software: Anaconda (all other software is installed within the Anaconda image)

Programming Language: Python

Source Code Address: https://github.com/qq492947833/DDL-MSPMF

# CRediT author statement

**Yuchen Ye**: Conceptualization, Formal analysis, Funding acquisition, Investigation, Methodology, Software, Visualization, Writing – Original Draft, Writing – Review & Editing. **Zixuan Qi:** Conceptualization, Data Curation, Formal analysis, Investigation, Methodology, Project administration, Resources, Supervision, Writing – Review & Editing. **Shixuan Li:** Data curation, Supervision, Writing – Review & Editing. **Wei Qi:** Data curation, Supervision, Writing – Review & Editing. **Yanpeng Cai:** Funding acquisition, Project administration, Resources, Supervision, Writing – Review & Editing. **Chaoxia Yuan:** Funding acquisition, Project administration, Resources, Supervision, Writing – Review & Editing.

# Acknowledgments

This research was supported by the National Natural Science Foundation of China (52439005), National Natural Science Foundation of China (42088101) and Postgraduate Research & Practice Innovation Program of Jiangsu Province (KYCX25_1593).

# References

Ali, H., Fowler, H. J., & Mishra, V. (2018). Global Observational Evidence of Strong Linkage Between Dew Point Temperature and Precipitation Extremes. *Geophysical Research Letters*, *45*(22), 12,320–312,330. https://doi.org/https://doi.org/10.1029/2018GL080557

Allen, M. R., & Ingram, W. J. (2002). Constraints on future changes in climate and the hydrologic cycle. *Nature*, *419*(6903), 224–232. https://doi.org/10.1038/nature01092

Ardabili, S., Mosavi, A., Dehghani, M., & Várkonyi-Kóczy, A. R. (2020). Deep learning and machine learning in hydrological processes climate change and earth systems a systematic review. *Engineering for Sustainable Future*, 52–62. https://doi.org/10.1007/978-3-030-36841-8_5

Argüeso, D., Evans, J. P., & Fita, L. (2013). Precipitation bias correction of very high resolution regional climate models. *Hydrol. Earth Syst. Sci.*, *17*(11), 4379–4388. https://doi.org/10.5194/hess-17-4379-2013

Ashouri, H., Hsu, K.-L., Sorooshian, S., Braithwaite, D. K., Knapp, K. R., Cecil, L. D., et al. (2015). PERSIANN-CDR: Daily precipitation climate data record from multisatellite observations for hydrological and climate studies. *Bulletin of the American Meteorological Society*, *96*(1), 69–83. https://doi.org/10.1175/BAMS-D-13-00068.1

Assiri, M. E., & Qureshi, S. (2022). A multi-source data fusion method to improve the accuracy of precipitation products: A machine learning algorithm. *Remote Sensing*, *14*(24), 6389. https://doi.org/10.3390/rs14246389

Awadallah, A. G., & Awadallah, N. A. (2013). A novel approach for the joint use of rainfall monthly and daily ground station data with TRMM data to generate IDF estimates in a poorly gauged arid region. *Open Journal of Modern Hydrology*, *3*(1), 1–7. https://doi.org/10.4236/ojmh.2013.31001

Bartsotas, N. S., Anagnostou, E. N., Nikolopoulos, E. I., & Kallos, G. (2018). Investigating satellite precipitation uncertainty over complex terrain. *Journal of Geophysical Research: Atmospheres*, *123*(10), 5346–5359. https://doi.org/10.1029/2017JD027559

Beck, H. E., Wood, E. F., Pan, M., Fisher, C. K., Miralles, D. G., Van Dijk, A. I. J. M., et al. (2019). MSWEP V2 Global 3-Hourly 0.1° precipitation: Methodology and quantitative assessment. *Bulletin of the American Meteorological Society*, *100*(3), 473–500. https://doi.org/10.1175/BAMS-D-17-0138.1

Bhatti, H. A., Rientjes, T., Haile, A. T., Habib, E., & Verhoef, W. (2016). Evaluation of bias correction method for satellite-based rainfall data. *Sensors*, *16*(6), 884. https://doi.org/10.3390/s16060884

Bi, K., Xie, L., Zhang, H., Chen, X., Gu, X., & Tian, Q. (2023). Accurate medium-range global weather forecasting with 3D neural networks. *Nature*, *619*(7970), 533–538. https://doi.org/10.1038/s41586-023-06185-3

Brocca, L., Hasenauer, S., Lacava, T., Melone, F., Moramarco, T., Wagner, W., et al. (2011). Soil moisture estimation through ASCAT and AMSR-E sensors: An intercomparison and validation study across Europe. *Remote Sensing of Environment*, *115*(12), 3390–3408. https://doi.org/10.1016/j.rse.2011.08.003

Carpenter, C. (2024). Physics-informed deep-learning models improve forecast scalability, reliability. *Journal of Petroleum Technology*, *76*(10), 90–93. https://doi.org/10.2118/1024-0090-JPT

Chen, C., Hao, J., Yang, S., & Li, Y. (2025a). Blending daily satellite precipitation product and rain gauges using stacking ensemble machine learning with the consideration of spatial heterogeneity. *Journal of Hydrology*, *658*, 133223. https://doi.org/10.1016/j.jhydrol.2025.133223

Chen, F. R., & Li, X. (2016). Evaluation of IMERG and TRMM 3B43 monthly precipitation products over Mainland China. *Remote Sensing*, *8*(6), 472. https://doi.org/10.3390/rs8060472

Chen, F. R., Wang, Y. G., & Li, X. (2024a). A data- and knowledge-driven method for fusing satellite-derived and ground-based precipitation observations. *IEEE Transactions on Geoscience and Remote Sensing*, *62*, 1–13. https://doi.org/10.1109/TGRS.2024.3385647

Chen, H. Q., Wen, D. B., Du, Y. N., Xiong, L. Y., & Wang, L. Y. (2023a). Errors of five satellite precipitation products for different rainfall intensities. *Atmospheric Research*, *285*, 106622. https://doi.org/10.1016/j.atmosres.2023.106622

Chen, J., Mei, J., Li, X., Lu, Y., Yu, Q., Wei, Q., et al. (2024b). TransUNet: Rethinking the U-Net architecture design for medical image segmentation through the lens of transformers. *Medical Image Analysis*, *97*, 103280. https://doi.org/10.1016/j.media.2024.103280

Chen, J. N., Lu, Y. Y., Yu, Q. H., Luo, X. D., Adeli, E., Wang, Y., et al. (2021). Transunet: Transformers make strong encoders for medical image segmentation. *ArXiv*, 04306. https://doi.org/10.48550/arXiv.2102.04306

Chen, L., Zhong, X., Zhang, F., Cheng, Y., Xu, Y., Qi, Y., et al. (2023b). FuXi: a cascade machine learning forecasting system for 15-day global weather forecast. *npj Climate and Atmospheric Science*, *6*(1), 190. https://doi.org/10.1038/s41612-023-00512-1

Chen, L. Z., Brun, P., Buri, P., Fatichi, S., Gessler, A., Mccarthy, M. J., et al. (2025b). Global increase in the occurrence and impact of multiyear droughts. *Science*, *387*(6731), 278–284. https://doi.org/10.1126/science.ado4245

Chen, T., & Guestrin, C. (2016). XGBoost: A Scalable Tree Boosting System. paper presented at Proceedings of the 22nd ACM SIGKDD International Conference on Knowledge Discovery and Data Mining, ACM, 2016/august. https://doi.org/doi

Chen, Z. Q., Qin, Y. X., Shen, Y., & Zhang, S. P. (2016). Evaluation of global satellite mapping of precipitation project daily precipitation estimates over the Chinese Mainland. *Advances in Meteorology*, *2016*(1), 9365294. https://doi.org/10.1155/2016/9365294

Copernicus Climate Change Service, & Climate Data Store (2023). ERA5 hourly data on single levels from 1940 to present. https://doi.org/10.24381/cds.adbb2d47

Deng, Q., Lu, P., Zhao, S., & Yuan, N. (2023). U-Net: A deep-learning method for improving summer precipitation forecasts in China. *Atmospheric and Oceanic Science Letters*, *16*(4), 100322. https://doi.org/10.1016/j.aosl.2022.100322

Deng, Y., Wang, X., Ruan, H. X., Lin, J. B., Chen, X. W., Chen, Y. N., et al. (2024). The magnitude and frequency of detected precipitation determine the accuracy performance of precipitation data sets in the high mountains of Asia. *Scientific Reports*, *14*(1). https://doi.org/10.1038/s41598-024-67665-8

Eekhout, J. P. C., Hunink, J. E., Terink, W., & De Vente, J. (2018). Why increased extreme precipitation under climate change negatively affects water security. *Hydrology and Earth System Sciences*, *22*(11), 5935–5946. https://doi.org/10.5194/hess-22-5935-2018

Feng, X., Porporato, A., & Rodriguez-Iturbe, I. (2013). Changes in rainfall seasonality in the tropics. *Nature Climate Change*, *3*(9), 811–815. https://doi.org/10.1038/nclimate1907

Gao, Y., Guan, J., Zhang, F., Wang, X., & Long, Z. (2022). Attention-Unet-Based Near-Real-Time Precipitation Estimation from Fengyun-4A Satellite Imageries. *Remote Sensing*, *14*(12). https://doi.org/10.3390/rs14122925

Gavahi, K., Foroumandi, E., & Moradkhani, H. (2023). A deep learning-based framework for multi-source precipitation fusion. *Remote Sensing of Environment*, *295*, 113723. https://doi.org/10.1016/j.rse.2023.113723

Green, A. C., Fowler, H. J., Blenkinsop, S., & Davies, P. A. (2025). Precipitation extremes in 2024. *Nature Reviews Earth & Environment*, *6*(4), 243–245. https://doi.org/10.1038/s43017-025-00666-x

Gu, J. X., Wang, Z. H., Kuen, J. S., Ma, L. Y., Shahroudy, A., Shuai, B., et al. (2018). Recent advances in convolutional neural networks. *Pattern Recognition*, *77*, 354–377. https://doi.org/10.1016/j.patcog.2017.10.013

Guo, L., Klingaman, N. P., Demory, M.-E., Vidale, P. L., Turner, A. G., & Stephan, C. C. (2018). The contributions of local and remote atmospheric moisture fluxes to East Asian precipitation and its variability. *Climate Dynamics*, *51*(11), 4139–4156. https://doi.org/10.1007/s00382-017-4064-4

Guo, W. F., Du, H., Guo, C., Southwell, B. J., Cheong, J. W., & Dempster, A. G. (2022). Information fusion for GNSS-R wind speed retrieval using statistically modified convolutional neural network. *Remote Sensing of Environment*, *272*, 112934. https://doi.org/10.1016/j.rse.2022.112934

Guo, Y., Li, Z. Y., Zhang, X., Chen, E. X., Bai, L. N., Tian, X., et al. (2012). Optimal Support Vector Machines for forest above-ground biomass estimation from multisource remote sensing data. paper presented at 2012 IEEE International Geoscience and Remote Sensing Symposium, IEEE, Munich, Germany, 2012/07. https://doi.org/doi

Han, J. Y., Miao, C. Y., Gou, J. J., Zheng, H. Y., Zhang, Q., & Guo, X. Y. (2023). A new daily gridded precipitation dataset for the Chinese mainland based on gauge observations. *Earth System Science Data*, *15*(7), 3147–3161. https://doi.org/10.5194/essd-15-3147-2023

Heo, J.-H., Ahn, H., Shin, J.-Y., Kjeldsen, T. R., & Jeong, C. (2019). Probability distributions for a quantile mapping technique for a bias correction of precipitation data: A case study to precipitation data under climate change. *Water*, *11*(7), 1475. https://doi.org/10.3390/w11071475

Hochreiter, S., & Schmidhuber, J. (1997). Long Short-term memory. *Neural Computation*, *9*(8), 1735–1780. https://doi.org/10.1162/neco.1997.9.8.1735

Hopfield, J. J. (1982). Neural networks and physical systems with emergent collective computational abilities. *Proceedings of the National Academy of Sciences*, *79*(8), 2554–2558. https://doi.org/10.1073/pnas.79.8.2554

Huffman, G. J., Stocker, E. F., Bolvin, D. T., Nelkin, E. J., & Tan, J. (2024). GPM IMERG Final Precipitation L3 1 day 0.1 degree x 0.1 degree V07. https://doi.org/10.5065/7DE2-M746

Jiang, Y. Z., Yang, K., Qi, Y. C., Zhou, X., He, J., Lu, H., et al. (2023). TPHiPr: a long-term (1979–2020) high-accuracy precipitation dataset (1/30°, daily) for the Third Pole region based on high-resolution atmospheric modeling and dense observations. *Earth System Science Data*, *15*(2), 621–638. https://doi.org/10.5194/essd-15-621-2023

Karpov, P., Godin, G., & Tetko, I. V. (2020). Transformer-CNN: Swiss knife for QSAR modeling and interpretation. *Journal of Cheminformatics*, *12*(1), 17. https://doi.org/10.1186/s13321-020-00423-w

Katiraie-Boroujerdy, P.-S., Rahnamay Naeini, M., Akbari Asanjan, A., Chavoshian, A., Hsu, K.-L., & Sorooshian, S. (2020). Bias correction of satellite-based precipitation estimations using quantile mapping approach in different climate regions of iran. *Remote Sensing*, *12*(13), 2102. https://doi.org/10.3390/rs12132102

Konapala, G., Mishra, A. K., Wada, Y., & Mann, M. E. (2020). Climate change will affect global water availability through compounding changes in seasonal precipitation and evaporation. *Nature Communications*, *11*(1), 3044. https://doi.org/10.1038/s41467-020-16757-w

Kotz, M., Levermann, A., & Wenz, L. (2022). The effect of rainfall changes on economic production. *Nature*, *601*(7892), 223–227. https://doi.org/10.1038/s41586-021-04283-8

Kubota, T., Aonashi, K., Ushio, T., Shige, S., Takayabu, Y. N., Kachi, M., et al. (2020). Global Satellite Mapping of Precipitation (GSMaP) products in the GPM Era, in *Satellite Precipitation Measurement: Volume 1*, edited by Levizzani, V., Kidd, C., Kirschbaum, D. B., Kummerow, C. D., Nakamura, K., & Turk, F. J., pp. 355–373, Springer International Publishing, Cham. https://doi.org/doi

Le, X.-H., Nguyen Van, L., Hai Nguyen, D., Nguyen, G. V., Jung, S., & Lee, G. (2023). Comparison of bias-corrected multisatellite precipitation products by deep learning framework. *International Journal of Applied Earth Observation and Geoinformation*, *116*, 103177. https://doi.org/10.1016/j.jag.2022.103177

Lecun, Y., Bottou, L., Bengio, Y., & Haffner, P. (1998). Gradient-based learning applied to document recognition. *Proceedings of the IEEE*, *86*(11), 2278–2324. https://doi.org/10.1109/5.726791

Lei, H. J., Zhao, H. Y., & Ao, T. Q. (2022a). A two-step merging strategy for incorporating multi-source precipitation products and gauge observations using machine learning classification and regression over China. *Hydrology and Earth System Sciences*, *26*(11), 2969–2995. https://doi.org/10.5194/hess-26-2969-2022

Lei, X. Y., Xu, W. L., Chen, S. T., Yu, T. T., Hu, Z. Y., Zhang, M., et al. (2022b). How well does the ERA5 reanalysis capture the extreme climate events over China? Part I: Extreme precipitation. *Frontiers in Environmental Science*, *10*, 921658. https://doi.org/10.3389/fenvs.2022.921658

Li, L. J., Wang, Y. T., Wang, L. Z., Hu, Q. F., Zhu, Z. D., Li, L. P., et al. (2022a). Spatio-temporal accuracy evaluation of MSWEP daily precipitation over the Huaihe River Basin, China: A comparison study with representative satellite- and reanalysis-based products. *Journal of Geographical Sciences*, *32*(11), 2271–2290. https://doi.org/10.1007/s11442-022-2047-9

Li, S., Chen, Y. N., Wei, W., Fang, G. H., & Duan, W. L. (2024a). The increase in extreme precipitation and its proportion over global land. *Journal of Hydrology*, *628*, 130456. https://doi.org/10.1016/j.jhydrol.2023.130456

Li, W., Chen, H., & Han, L. (2024b). Improving Explainability of Deep Learning for Polarimetric Radar Rainfall Estimation. *Geophysical Research Letters*, *51*(11), e2023GL107898. https://doi.org/10.1029/2023GL107898

Li, W. Y., Jiang, Q., He, X. G., Sun, H. Q., Sun, W. W., Scaioni, M., et al. (2022b). Effective multi-satellite precipitation fusion procedure conditioned by gauge background fields over the Chinese mainland. *Journal of Hydrology*, *610*, 127783. https://doi.org/10.1016/j.jhydrol.2022.127783

Li, Y., Wang, W. S., Wang, G. Q., & Yu, S. Y. (2021). Evaluation and hydrological application of a data fusing method of multi-source precipitation products-a case study over Tuojiang River Basin. *Remote Sensing*, *13*(13), 2630. https://doi.org/10.3390/rs13132630

Lundberg, S. M., & Lee, S.-I. (2017). A unified approach to interpreting model predictions. paper presented at Proceedings of the 31st International Conference on Neural Information Processing Systems, Curran Associates Inc., Long Beach, California, USA. https://doi.org/doi

Luo, Y., Xu, X., Liu, Y., Chao, H., Chu, H., Chen, L., et al. (2022). Robust Precipitation Bias Correction Through an Ordinal Distribution Autoencoder. *IEEE Intelligent Systems*, *37*(1), 60–70. https://doi.org/10.1109/MIS.2021.3088543

Lyu, Y., & Yong, B. (2024). A novel Double Machine Learning strategy for producing high‐precision multi‐source merging precipitation estimates over the Tibetan Plateau. *Water Resources Research*, *60*(4), e2023WR035643. https://doi.org/10.1029/2023WR035643

Lyu, Y., & Yong, B. (2025). Using an Explainable Machine Learning Approach to Produce High-Resolution Hourly Precipitation Estimates for a Typical Data-Deficiency Basin. *Journal of Geophysical Research: Machine Learning and Computation*, *2*(1), e2024JH000489. https://doi.org/10.1029/2024JH000489

Maggioni, V., Massari, C., & Kidd, C. (2022). Errors and uncertainties associated with quasiglobal satellite precipitation products. *Precipitation Science*, 377–390. https://doi.org/10.1016/B978-0-12-822973-6.00023-8

Malik, K., Robertson, C., Braun, D., & Greig, C. (2021). U-Net convolutional neural network models for detecting and quantifying placer mining disturbances at watershed scales. *International Journal of Applied Earth Observation and Geoinformation*, *104*, 102510. https://doi.org/10.1016/j.jag.2021.102510

Massari, C., Brocca, L., Pellarin, T., Abramowitz, G., Filippucci, P., Ciabatta, L., et al. (2020). A daily 25 km short-latency rainfall product for data-scarce regions based on the integration of the Global Precipitation Measurement mission rainfall and multiple-satellite soil moisture products. *Hydrology and Earth System Sciences*, *24*(5), 2687–2710. https://doi.org/10.5194/hess-24-2687-2020

Mendez, M., Maathuis, B., Hein-Griggs, D., & Alvarado-Gamboa, L.-F. (2020). Performance evaluation of bias correction methods for climate change monthly precipitation projections over Costa Rica. *Water*, *12*(2), 482. https://doi.org/10.3390/w12020482

Miao, C., Ashouri, H., Hsu, K.-L., Sorooshian, S., & Duan, Q. (2015). Evaluation of the PERSIANN-CDR daily rainfall estimates in capturing the behavior of extreme precipitation events over China. *Journal of Hydrometeorology*, *16*(3), 1387–1396. https://doi.org/10.1175/JHM-D-14-0174.1

Mu, B., Qin, B., Yuan, S. J., Wang, X., & Chen, Y. X. (2023). PIRT: A physics-informed red tide deep learning forecast model considering causal-inferred predictors selection. *IEEE Geoscience and Remote Sensing Letters*, *20*, 1–5. https://doi.org/10.1109/LGRS.2023.3250642

Myhre, G., Alterskjær, K., Stjern, C. W., Hodnebrog, Ø., Marelle, L., Samset, B. H., et al. (2019). Frequency of extreme precipitation increases extensively with event rareness under global warming. *Scientific Reports*, *9*(1), 16063. https://doi.org/10.1038/s41598-019-52277-4

Nan, L. J., Yang, M. X., Wang, H., Wang, H. J., & Dong, N. P. (2024). An innovative correction–fusion approach for multi-satellite precipitation products conditioned by gauge background fields over the Lancang River Basin. *Remote Sensing*, *16*(11), 1824. https://doi.org/10.3390/rs16111824

Newman, R., & Noy, I. (2023). The global costs of extreme weather that are attributable to climate change. *Nature Communications*, *14*(1), 6103. https://doi.org/10.1038/s41467-023-41888-1

Nguyen, G. V., Le, X.-H., Van, L. N., Jung, S., Yeon, M., & Lee, G. (2021). Application of random forest algorithm for merging multiple satellite precipitation products across South Korea. *Remote Sensing*, *13*(20), 4033. https://doi.org/10.3390/rs13204033

Nie, S. P., Luo, Y., Wu, T. W., Shi, X. L., & Wang, Z. Z. (2015). A merging scheme for constructing daily precipitation analyses based on objective bias‐correction and error estimation techniques. *Journal of Geophysical Research: Atmospheres*, *120*(17), 8671‐8692. https://doi.org/10.1002/2015JD023347

Noaa National Centers for Environmental Information (2019). OAA Climate Data Record (CDR) of CPC Morphing Technique (CMORPH) High Resolution Global Precipitation Estimates, Version 1. . https://doi.org/10.25921/w9va-q159

Owens, R. G., & Hewson, T. D. (2018). ECMWF forecast user guide. https://doi.org/10.21957/m1cs7h

Pan, Y., Yuan, Q. Q., Ma, J. S., & Wang, L. C. (2022). Improved daily spatial precipitation estimation by merging multi-source precipitation data based on the geographically weighted regression method: A case study of Taihu Lake Basin, China. *International Journal of Environmental Research and Public Health*, *19*(21), 13866. https://doi.org/10.3390/ijerph192113866

Pizzorni, M., Innocenti, A., & Tollin, N. (2024). Droughts and floods in a changing climate and implications for multi-hazard urban planning: A review. *City and Environment Interactions*, *24*, 100169. https://doi.org/10.1016/j.cacint.2024.100169

Prein, A. F., & Gobiet, A. (2016). Impacts of uncertainties in European gridded precipitation observations on regional climate analysis. *International Journal of Climatology*, *37*(1), 305–327. https://doi.org/10.1002/joc.4706

Qi, S. S., Lv, A. F., Wang, G. S., & Zhang, C. H. (2023). A Multiplicative-Exponential function to correct precipitation for distributed hydrological modeling in Poorly-gauged basins. *Journal of Hydrology*, *620*, 129393. https://doi.org/10.1016/j.jhydrol.2023.129393

Qi, Y., Zhang, J., Cao, Q., Hong, Y., & Hu, X.-M. (2013). Correction of Radar QPE Errors for Nonuniform VPRs in Mesoscale Convective Systems Using TRMM Observations. *Journal of Hydrometeorology*, *14*(5), 1672–1682. https://doi.org/10.1175/JHM-D-12-0165.1

Quinlan, J. R. (1979). Discovering rules from large collections of examples: Acase study. *Expert Systems in the Microelectronic Age*, *2*(9), 168–201.

Reichle, R. H., & Koster, R. D. (2004). Bias reduction in short records of satellite soil moisture. *Geophysical Research Letters*, *31*(19), 2–5. https://doi.org/10.1029/2004GL020938

Rhodes, R. I., Shaffrey, L. C., & Gray, S. L. (2014). Can reanalyses represent extreme precipitation over England and Wales? *Quarterly Journal of the Royal Meteorological Society*, *141*(689), 1114–1120. https://doi.org/10.1002/qj.2418

Ronneberger, O., Fischer, P., & Brox, T. (2015). U-Net: Convolutional networks for biomedical image segmentation. paper presented at Medical Image Computing and Computer-Assisted Intervention – MICCAI 2015, Springer International Publishing, Cham, 2015//. https://doi.org/doi

Sebastianelli, S., Russo, F., Napolitano, F., & Baldini, L. (2013). On precipitation measurements collected by a weather radar and a rain gauge network. *Nat. Hazards Earth Syst. Sci.*, *13*(3), 605–623. https://doi.org/10.5194/nhess-13-605-2013

Senocak, A. U. G., Yilmaz, M. T., Kalkan, S., Yucel, I., & Amjad, M. (2023). An explainable two-stage machine learning approach for precipitation forecast. *Journal of Hydrology*, *627*, 130375. https://doi.org/10.1016/j.jhydrol.2023.130375

Shen, L. L., Lin, R. S., Lu, L., Xu, C., & Liu, Y. (2020). Accuracy analysis of IMERG and CMORPH precipitation data over North China. *Climate Research*, *81*, 55–70. https://doi.org/10.3354/cr01610

Sun, J., Yang, K., He, X., Wang, G., Wang, Y., Yu, Y., et al. (2025). Causal pathways underlying global soil moisture–precipitation coupling. *Nature Communications*, *16*(1), 8935. https://doi.org/10.1038/s41467-025-63999-7

Sun, X. M., & Barros, A. P. (2010). An evaluation of the statistics of rainfall extremes in rain gauge observations, and satellite-based and reanalysis products using universal multifractals. *Journal of Hydrometeorology*, *11*(2), 388–404. https://doi.org/10.1175/2009JHM1142.1

Tabari, H. (2020). Climate change impact on flood and extreme precipitation increases with water availability. *Scientific Reports*, *10*(1), 13768. https://doi.org/10.1038/s41598-020-70816-2

Van Der Ent, R. J., & Savenije, H. H. G. (2011). Length and time scales of atmospheric moisture recycling. *Atmos. Chem. Phys.*, *11*(5), 1853–1863. https://doi.org/10.5194/acp-11-1853-2011

Vaswani, A., Shazeer, N., Parmar, N., Uszkoreit, J., Jones, L., Gomez, A. N., et al. (2017). Attention is all you need. paper presented at Proceedings of the 31st International Conference on Neural Information Processing Systems, Curran Associates Inc., Long Beach, California, USA. https://doi.org/doi

Wang, C.-C. (2014). On the calculation and correction of equitable threat score for model quantitative precipitation forecasts for small verification areas: The example of

Taiwan. *Weather and Forecasting*, *29*(4), 788–798. https://doi.org/10.1175/WAF-D-13-00087.1

Wang, D., Yang, S., Li, X., Peng, J., Ma, H., & Wu, X. (2024). Multiscale Attention-UNet-Based Near-Real-Time Precipitation Estimation From FY-4A/AGRI and Doppler Radar Observations. *IEEE Journal of Selected Topics in Applied Earth Observations and Remote Sensing*, *17*, 19998–20011. https://doi.org/10.1109/jstars.2024.3488854

Wang, X., Shi, S. Y., Zhu, L. T., Nie, Y. F., & Lai, G. J. (2023). Traditional and novel methods of rainfall observation and measurement: A review. *Journal of Hydrometeorology*, *24*(12), 2153–2176. https://doi.org/10.1175/JHM-D-22-0122.1

Wilson, D. R., & Martinez, T. R. (Year). The need for small learning rates on large problems. paper presented at IJCNN'01. International Joint Conference on Neural Networks. Proceedings (Cat. No.01CH37222), 15–19 July 2001. https://doi.org/doi

Wu, J., & Gao, X. J. (2013). A gridded daily observation dataset over China region and comparison with the other datasets. *Chinese Journal of Geophysics*, *56*(4), 1102–1111. https://doi.org/10.6038/cjg20130406

Wu, Y. P., Yin, X. W., Zhou, G. Y., Bruijnzeel, L. A., Dai, A. G., Wang, F., et al. (2024). Rising rainfall intensity induces spatially divergent hydrological changes within a large river basin. *Nature Communications*, *15*(1), 823. https://doi.org/10.1038/s41467-023-44562-8

Xu, X. Z., Sun, X. Y., Han, W., Zhong, X. H., Chen, L., Gao, Z. Q., et al. (2025). FuXi-DA: A generalized deep learning data assimilation framework for assimilating satellite observations. *npj Climate and Atmospheric Science*, *8*(1), 156. https://doi.org/10.1038/s41612-025-01039-3

Xu, Y., Tang, G. Q., Li, L. J., & Wan, W. (2024). Multi-source precipitation estimation using machine learning: Clarification and benchmarking. *Journal of Hydrology*, *635*, 131195. https://doi.org/10.1016/j.jhydrol.2024.131195

Yang, X. Y., Yang, S., Tan, M. L., Pan, H. Y., Zhang, H. L., Wang, G. Q., et al. (2022). Correcting the bias of daily satellite precipitation estimates in tropical regions using deep neural network. *Journal of Hydrology*, *608*, 127656. https://doi.org/10.1016/j.jhydrol.2022.127656

You, Y., & Ting, M. (2021). Low Pressure Systems and Extreme Precipitation in Southeast and East Asian Monsoon Regions. *Journal of Climate*, *34*(3), 1147–1162. https://doi.org/10.1175/JCLI-D-20-0206.1

Yu, C., Hu, D. Y., Liu, M. Q., Wang, S. S., & Di, Y. F. (2020). Spatio-temporal accuracy evaluation of three high-resolution satellite precipitation products in China area. *Atmospheric Research*, *241*, 104952. https://doi.org/10.1016/j.atmosres.2020.104952

Yuan, S., Zhu, S., Luo, X., & Mu, B. (2024). A deep learning-based bias correction model for Arctic sea ice concentration towards MITgcm. *Ocean Modelling*, *188*, 102326. https://doi.org/10.1016/j.ocemod.2024.102326

Zhai, L. P., & Lee, J. E. (2023). Harnessing ICT resources to enhance community disaster resilience: A case study of employing social media to Zhengzhou 7.20 Rainstorm, China. *Water*, *15*(19), 3516. https://doi.org/10.3390/w15193516

Zhang, J. M., Xu, J. H., Dai, X. A., Ruan, H. H., Liu, X. L., & Jing, W. L. (2022). Multi-source precipitation data merging for heavy rainfall events based on cokriging and

machine learning methods. *Remote Sensing*, *14*(7), 1750. https://doi.org/10.3390/rs14071750

Zhao, Y. M., Xu, K., Dong, N. P., & Wang, H. (2022a). Optimally integrating multi-source products for improving long series precipitation precision by using machine learning methods. *Journal of Hydrology*, *609*, 127707. https://doi.org/10.1016/j.jhydrol.2022.127707

Zhao, Y. Y., Zhang, J. H., Bai, Y., Zhang, S., Yang, S. S., Henchiri, M., et al. (2022b). Drought monitoring and performance evaluation based on machine learning fusion of multi-source remote sensing drought factors. *Remote Sensing*, *14*(24), 6398. https://doi.org/10.3390/rs14246398

Zhou, Y. T., Zhan, R. F., Wang, Y. Q., Chen, P. Y., Tan, Z. M., Xie, Z. P., et al. (2024). A physics-informed deep-learning intensity prediction scheme for tropical cyclones over the Western North Pacific. *Advances in Atmospheric Sciences*, *41*(7), 1391–1402. https://doi.org/10.1007/s00376-024-3282-z

Supplementary Information

# A Dual-TransUNet Deep Learning Framework for Multi-Source Precipitation Merging and Improving Seasonal and Extreme Estimates

Yuchen Ye[1,2], Zixuan Qi[4], Shixuan Li[4], Wei Qi[4], Yanpeng Cai[4], Chaoxia Yuan[1,3,*]

[1] State Key Laboratory of Climate System Prediction and Risk Management (CPRM) /Key Laboratory of Meteorological Disaster, Ministry of Education (KLME) /Collaborative Innovation Center on Forecast and Evaluation of Meteorological Disasters (CIC-FEMD), Nanjing University of Information Science and Technology, Nanjing 210044, China

[2] School of Atmospheric Sciences, Nanjing University of Information Science and Technology, Nanjing 210044, China

[3] School of Future Technology, Nanjing University of Information Science and Technology, Nanjing 210044, China

[4] Guangdong Basic Research Center of Excellence for Ecological Security and Green Development, Guangdong Provincial Key Laboratory of Water Quality Improvement and Ecological Restoration for Watersheds, School of Ecology, Environment and Resources, Guangdong University of Technology, Guangzhou, 510006, China

Corresponding author: Yuan Chaoxia (chaoxia.yuan@nuist.edu.cn)

**Contents of this file**

Tables S1-S2

Figures S1-S11

**Table S1.** Details of the dataset used in the study

| Dataset name | Temporal/spatial resolution | Time/Space Coverage | Dataset link | Institution |
|---|---|---|---|---|
| CMORPH | 1hour/0.25° | Global/1998-Today | https://www.ncei.noaa.gov/products/climate-data-records/precipitation-cmorph | NOAA |
| PERSIANN | 1day/0.25° | -60°S-60°N,-180°W-180°E/1983-2021 | https://climatedataguide.ucar.edu/climate-data/persiann-cdr-precipitation-estimation-remotely-sensed-information-using-artificial | NOAA |
| GPM | 1day/0.1° | Global/1998-Today | https://gpm.nasa.gov/ | NASA |
| GSMAP | 1hour/0.1° | -60°S-60°N,-180°W-180°E/2000-Today | https://sharaku.eorc.jaxa.jp/GSMaP/guide.html | JAXA |
| MSWEP | 3hour/0.1° | Global/1979-Today | https://www.gloh2o.org/mswep/ | GloH2O |
| ERA5 | 1hour/0.25° | Global/1940-Today | https://cds.climate.copernicus.eu/datasets/reanalysis-era5-single-levels | ECMWF |
| CN05.1 | 1day/0.25° | 69.75°E–140.25°E, 14.75°N–55.25°N/1961-Today | https://ccrc.iap.ac.cn/resource/detail?id=228 | CCRC |
| TPHiPr | 1day/(1/30)° | 61.03°E–105.66°E, 25.73°N–41.36°N/1979-2020 | https://data.tpdc.ac.cn/en/data/e45be858-bcb2-4fea-bd10-5c2662cb34a5 | TPDC |

**Table S2.** The spatial average R and RMSE between multi-source precipitation data fusion correction results from different deep learning models and CN05.1 observed precipitation, along with the total runtime of each model combination.

|  | R | RMSE | Total runtime |
|---|---|---|---|
| CNN Transformer + CNN Transformer | 0.3973 | 4.1591 | 100,000s |
| UNet + UNet | 0.7309 | 2.8445 | 11,000s |
| TransUNet + TransUNet | 0.7512 | 2.6954 | 14,000s |
| Transformer + Transformer | 0.6598 | 3.0251 | 566,000s |
| XGBoost + LSTM | 0.6268 | 19.8655 | 698,300s |
| TransUNet Direct | 0.7031 | 2.9401 | 7,000s |

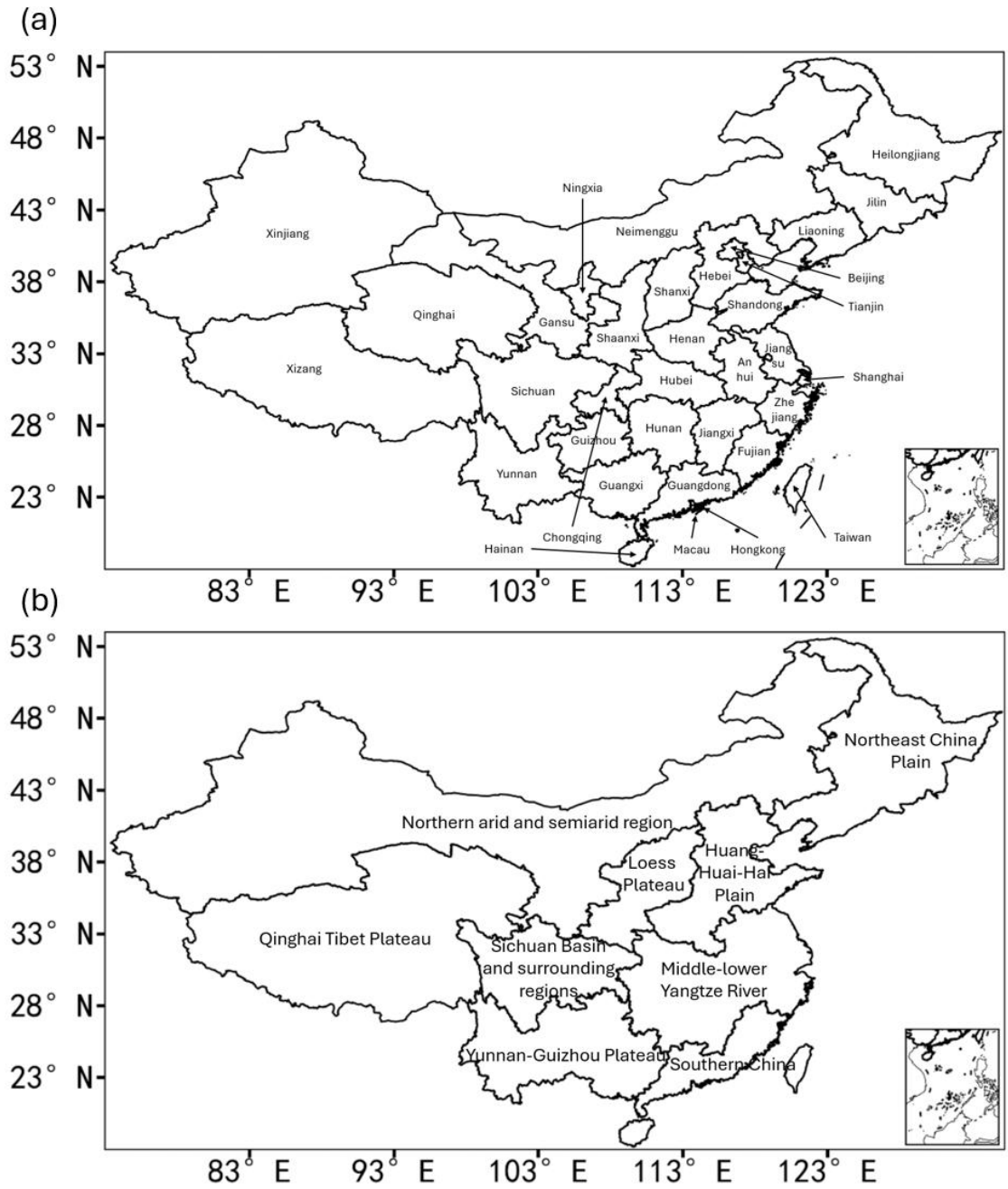

**Fig. S1.** Map of China's provincial administrative boundaries (a) and the boundaries of China's nine major agricultural regions (b); Text in this figure represents the name of each province / agricultural region.

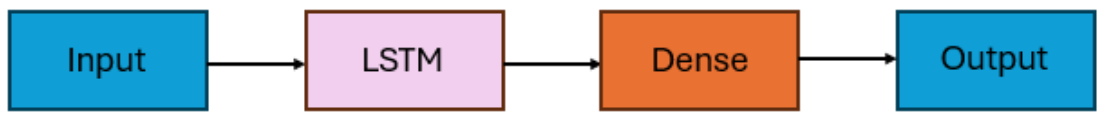

**Fig. S2.** The LSTM network architecture diagram used in this study.

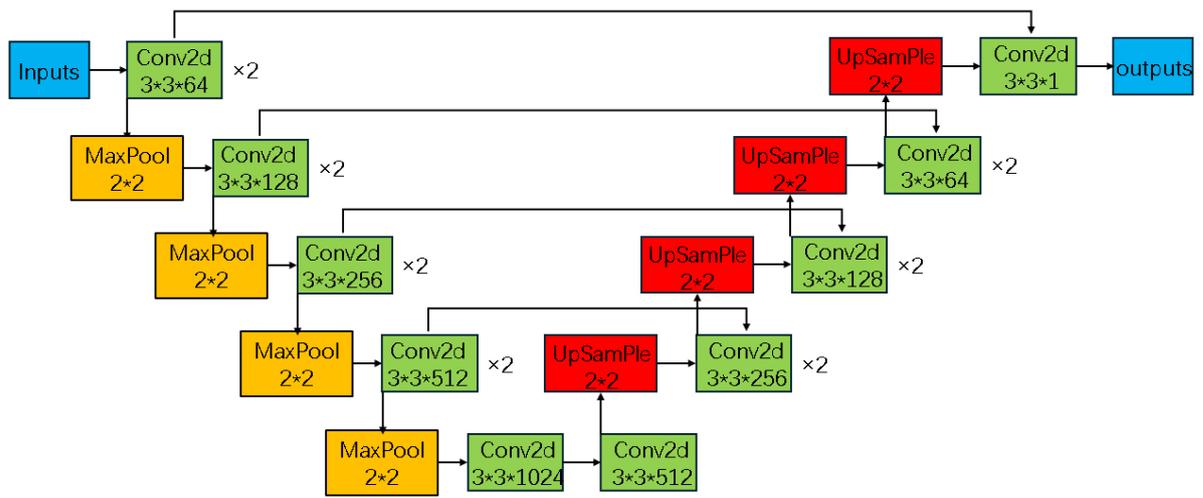

**Fig. S3.** The UNet network architecture diagram used in this study.

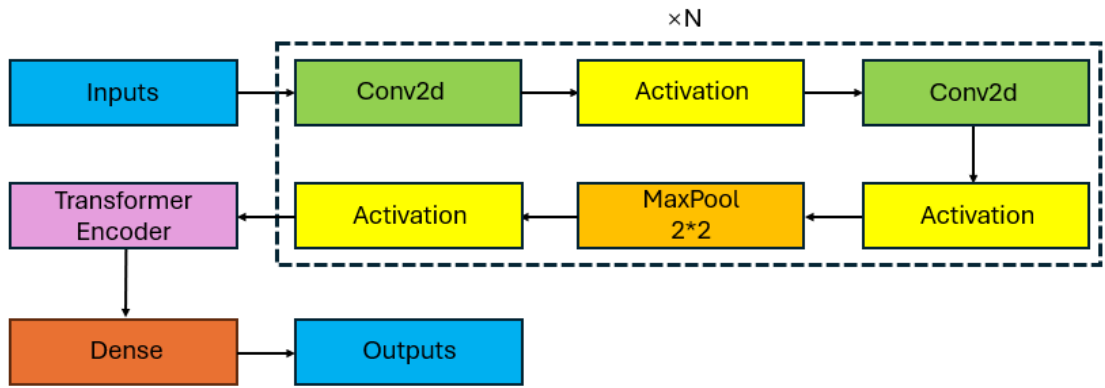

**Fig. S4.** The CNN Transformer network architecture diagram used in this study.

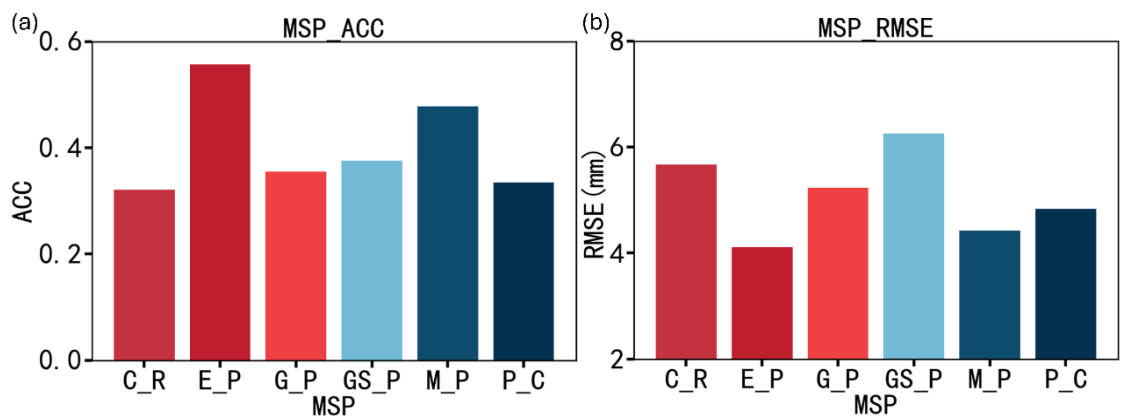

**Fig. S5.** Spatial average correlation coefficient (a) / root mean square error (b) between the MSPs data and the CN05.1 observed precipitation.

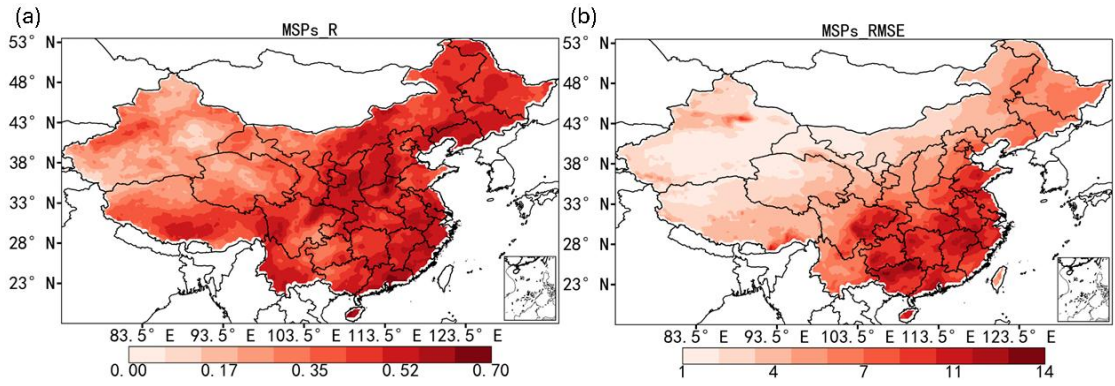

**Fig. S6.** The spatial maps of correlation coefficients (a) / root mean square errors (b) between the MSPs ensemble mean results and CN05.1 observed precipitation.

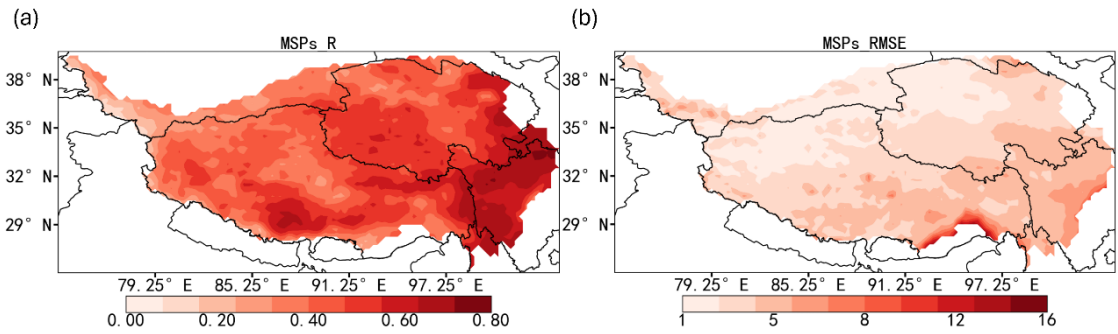

**Fig. S7.** The spatial maps of correlation coefficients (a) / root mean square errors (b) between the MSPs ensemble mean results and TPHiPr observed precipitation.

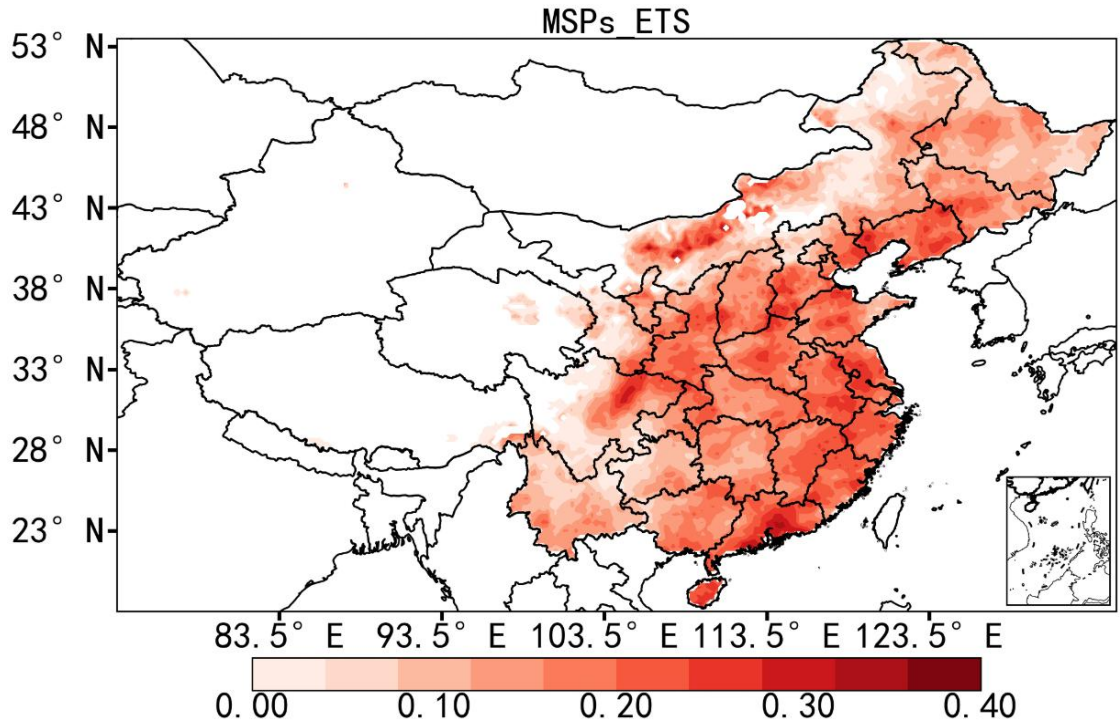

**Fig. S8.** Spatial map of ETS for identifying heavy precipitation events by MSPs ensemble mean; The white areas within China indicate regions without extreme precipitation events where daily rainfall exceeds 25mm.

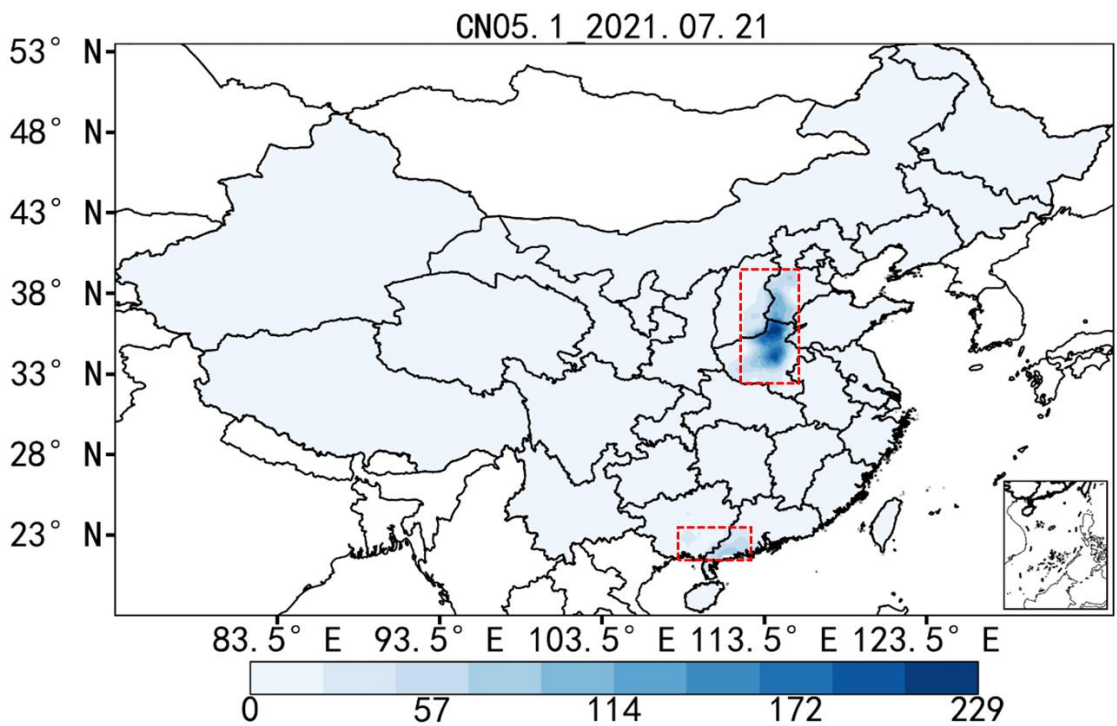

**Fig. S9.** CN05.1 data on the spatial distribution of rainfall in China on July 21, 2021; The unit of precipitation is mm/day.

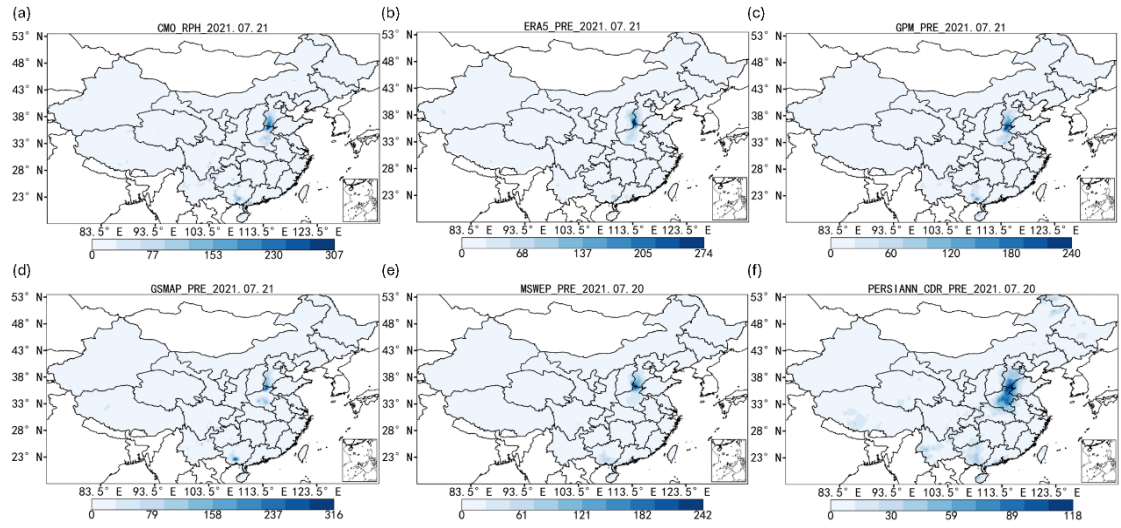

**Fig. S10.** MSPs data on the spatial distribution of rainfall in China on July 21, 2021; The unit of precipitation is mm/day.

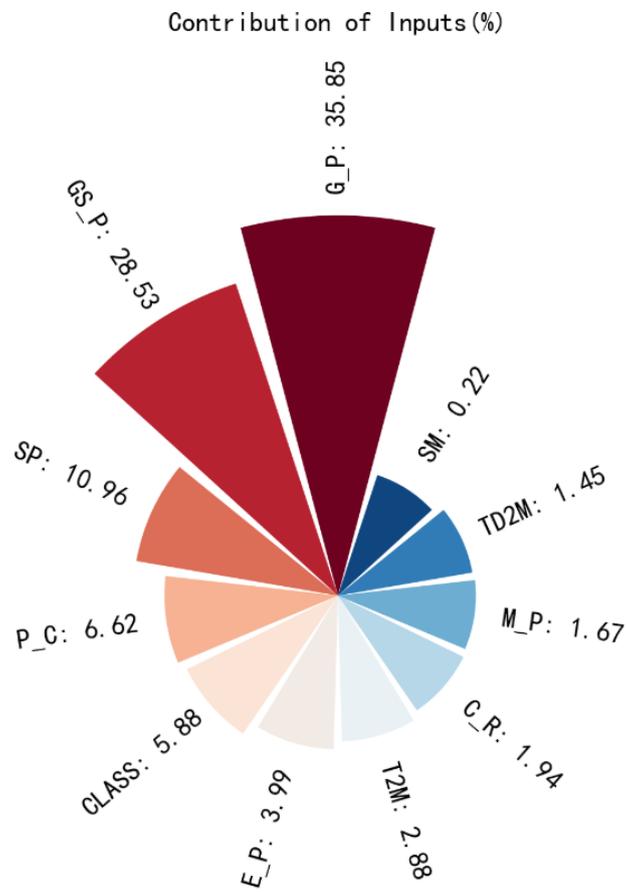

**Fig. S11.** Relative contributions of different forecast factors in the TransUNet model for MSPs fusion correction; The size of the sector is proportional to the magnitude of the predictor's contribution.